%% file: arxiv.tex
\documentclass{article}

\usepackage[preprint,nonatbib]{neurips_2022}

\usepackage[pdftex]{graphicx}

\usepackage[utf8]{inputenc} 
\usepackage[T1]{fontenc}    
\usepackage{hyperref}       
\hypersetup{colorlinks,linkcolor=blue, citecolor=red}
\usepackage{url}            
\usepackage{booktabs}       
\usepackage{amsfonts}       
\usepackage{nicefrac}       
\usepackage{microtype}      

\usepackage{xcolor}

\usepackage{amsmath}
\usepackage{amssymb}
\usepackage{bm}

\usepackage{float}
\usepackage{graphicx}
\usepackage{caption}
\usepackage{subcaption}
\usepackage{enumitem}

\usepackage{titlesec}      
\usepackage[hang,flushmargin]{footmisc}

\def\<{\langle} \def\>{\rangle}

\usepackage{amsthm}

\newcommand{\ls}[1]{}

\let\svthefootnote\thefootnote
\newcommand\blfootnote[1]{%
  \let\thefootnote\relax%
  \footnotetext{#1}%
  \let\thefootnote\svthefootnote%
}

\allowdisplaybreaks

\title{An Analytical Theory of Curriculum Learning in Teacher-Student Networks}

\author{%
Luca Saglietti$^{\dag,*}$, Stefano Sarao Mannelli$^{\ddag,*}$, and Andrew Saxe$^{\ddag,\S}$
}

\begin{document}

\maketitle
\blfootnote{\noindent
$\dag$ Department of Computing Sciences, Bocconi University. \\
$\ddag$ Gatsby Computational Neuroscience Unit \& Sainsbury Wellcome Centre, University College London. \\
$\S$ FAIR, Meta AI \\
$*$ Equal contributions.
}

\input{camera_ready_abstract}
\input{camera_ready_text}

\bibliography{bib}
\bibliographystyle{unsrt}

\newpage

\hrule height 0.1cm
\vspace{0.25in}
  \begin{center}
	{\huge \textbf{Supplemental Material}}
\end{center}
\vspace{0.29in}
\hrule height 0.025cm
\vspace{1cm}
\input{camera_ready_ap}

\end{document}

%% file: camera_ready_abstract.tex
\begin{abstract}
    In animals and humans, curriculum learning---presenting data in a curated order---is critical to rapid learning and effective pedagogy. 
    A long history of experiments has demonstrated the impact of curricula in a variety of animals but, despite its ubiquitous presence, a theoretical understanding of the phenomenon is still lacking. 
    Surprisingly, in contrast to animal learning, curricula strategies are not widely used in machine learning and recent simulation studies reach the conclusion that curricula are moderately effective or even ineffective in most cases. 
    This stark difference in the importance of curriculum raises a fundamental theoretical question: when and why does curriculum learning help? 
    In this work, we analyse a prototypical neural network model of curriculum learning in the high-dimensional limit, employing statistical physics methods. 
    We study a task in which a sparse set of informative features are embedded amidst a large set of noisy features. We analytically derive average learning trajectories for simple neural networks on this task, which establish a clear speed benefit for curriculum learning in the online setting. However, when training experiences can be stored and replayed 
    the advantage of curriculum in standard neural networks disappears, in line with observations from the deep learning literature. 
    Inspired by synaptic consolidation techniques developed to combat catastrophic forgetting, we propose curriculum-aware algorithms that consolidate synapses at curriculum change points and investigate whether this can boost the benefits of curricula. We derive generalisation performance as a function of consolidation strength (implemented as an $L_2$ regularisation/elastic coupling connecting learning phases), and show that curriculum-aware algorithms can yield a large improvement in test performance.
    Our reduced analytical descriptions help reconcile apparently conflicting empirical results, trace regimes where curriculum learning yields the largest gains, and provide experimentally-accessible predictions for the impact of task parameters on curriculum benefits. More broadly, our results suggest that fully exploiting a curriculum may require explicit adjustments in the loss.
\end{abstract}

%% file: camera_ready_text.tex
\section{Introduction}

Presenting learning materials in a meaningful order according to a curriculum greatly helps learning in animals and humans \cite{lawrence1952transfer,baker1954discrimination,elio_effects_1984,wilson_eighty_2019}, and is considered an essential aspect of good pedagogy \cite{avrahami_teaching_1997}. For example, humans have been shown to learn visual discriminations faster when presented with examples that exaggerate the relevant difference between classes, a phenomenon known as ``fading'' \cite{pashler_when_2013,hornsby_improved_2014,roads_easy--hard_2018}. Beyond humans, curricula in the form of ``shaping'' or ``staircase'' procedures are a near-universal feature of task designs in animal studies, without which training often fails entirely. 
For instance, the International Brain Laboratory task, a standardised perceptual decision-making training paradigm in mice, involves six stages of increasing difficulty before reaching final performance \cite{the_international_brain_laboratory_standardized_2021}.

Building from this intuition, a seminal series of papers proposed a similar curriculum learning approach for machine learning (ML) \cite{elman_learning_1993,krueger_flexible_2009,bengio2009curriculum}. In striking contrast to the clear benefits of curriculum in biological systems, however, curriculum learning has generally yielded equivocal benefits in artificial systems. Experiments in a variety of domains \cite{pentina_curriculum_2015,HacohenW19} have found usually modest speed and generalisation improvements from curricula. Recent extensive empirical analyses have found minimal benefits on standard datasets \cite{wu2020curricula}. Indeed, a common intuition in deep learning practice holds that training distributions should ideally be as close as possible to testing distributions, a notion which runs counter to curriculum. Perhaps the only areas where curricula are actively used are in large language models \cite{brown_language_2020} and certain reinforcement learning settings \cite{jiang_prioritized_2021}.

This gap between the effect of curriculum in biological and artificial learning systems poses a puzzle for theory. When and why is curriculum learning useful? What properties of a task determine the extent of possible benefits? What ordering of learning material is most beneficial? And can new learning algorithms better exploit curricula? Compared to the empirical investigations of curriculum learning, theoretical results on curriculum learning remain sparse. Most notably, \cite{weinshall2018curriculum,weinshall2020theory} show that curriculum can lead to faster learning in a simple setting, but the effects of curriculum on asymptotic generalisation and the dependence on task structure remain unclear.
A hint that indeed curriculum learning might lead to statistically different minima comes from a connection between constraint-satisfaction problems and physics results on flow networks \cite{ruiz2019tuning}, but to our knowledge no direct result has been reported in the modern theoretical ML literature. 

In this work we study the impact of curriculum using the analytically tractable teacher-student framework and the tools of statistical physics \cite{mezard1987spin,engel2001statistical,zdeborova2016statistical,bahri2020statistical}. High-dimensional teacher-student models are a popular approach for systematically studying learning behaviour in neural networks \cite{cugliandolo1993analytical,biehl1995learning,engel2001statistical}, and have recently been leveraged to analyse a variety of phenomena \cite{advani2020highdimensional,goldt2019dynamics,mannelli2019passed,mannelli2019afraid,mannelli2020complex,cui2020large}.
Using a simple model to build structured data \cite{bengio2009curriculum}, we examine the impact of ordering examples by increasing difficulty (curriculum), decreasing difficulty (anti-curriculum), or standard shuffled training. We derive exact expressions for the online learning dynamics and the performance of batch learning. However, in the latter, curriculum confers no benefit under standard training in our model setting. Motivated by theories of synaptic consolidation and elastic weight consolidation \cite{zenke2017continual,kirkpatrick2017overcoming}, we introduce elastic penalties (Gaussian priors) that regularise training toward solutions obtained in prior curriculum phases, instantiating a long-term memory effect. With these priors, curriculum yields benefits both in the online \ref{sec:online_dyn} and in the batch \ref{sec:batch} settings. 

\vspace{-0.2cm}
\paragraph*{Further related work.} 
The first empirical investigation of curriculum learning appeared in 1927 ~\cite{pavlov2010conditioned}, consisting in a visual discrimination task for dogs under curriculum and no-curriculum paradigms. Later behavioural studies proved curricula to be beneficial independent of the animal (dogs, mice, rats, pigeons, humans) and the data modality (visual, auditory, or tactile stimuli) \cite{skinner2019behavior,lawrence1952transfer,baker1954discrimination,ahissar1997task,morris1977levels,pashler_when_2013}. 
However, these experimental observations were not observed in standard artificial neural networks (ANNs). Several ideas in the connectionist community were proposed in order to show curriculum effects in the learning dynamics of ANNs~\cite{plunkett1991rote,plunkett1991u,elman_learning_1993,krueger_flexible_2009}. While these studies were able to match previous experimental data, they also required substantial changes in the architecture of the ANN and/or in the learning rule.

Except for very few instances \cite{brown_language_2020,jiang_prioritized_2021}, standard ML practice tends to avoid taking curricula into account. An obvious obstacle is the fact that most datasets do not provide meta-data about sample difficulties. An interesting line of research pointed out the possible relevance of implicit curricula, based on the observation that neural networks tend to consistently learn the samples in a certain order \cite{toneva2019empirical}. Thus, a possible way of addressing the lack of difficulty labels would be to use the natural learning order as indicative of the various difficulties of the training samples. However, a recent work~\cite{wu2020curricula}, which compared several heuristics for curriculum learning ---including implicit curricula--- in a variety of settings, showed limited benefits with this strategy. 

The picture that emerges from the literature seems contradictory: on the one hand, curricula appear fundamental to biological learning; on the other hand, curricula appear largely irrelevant in many machine learning settings. The core motivation behind our work is to reconcile these views and contribute to a theoretical understanding of curriculum learning. 

\section{Model definition and overview of approach}\label{sec:problem}

\begin{figure}
	\centering
    \includegraphics[trim={0 0 0 0},clip,width=\linewidth]{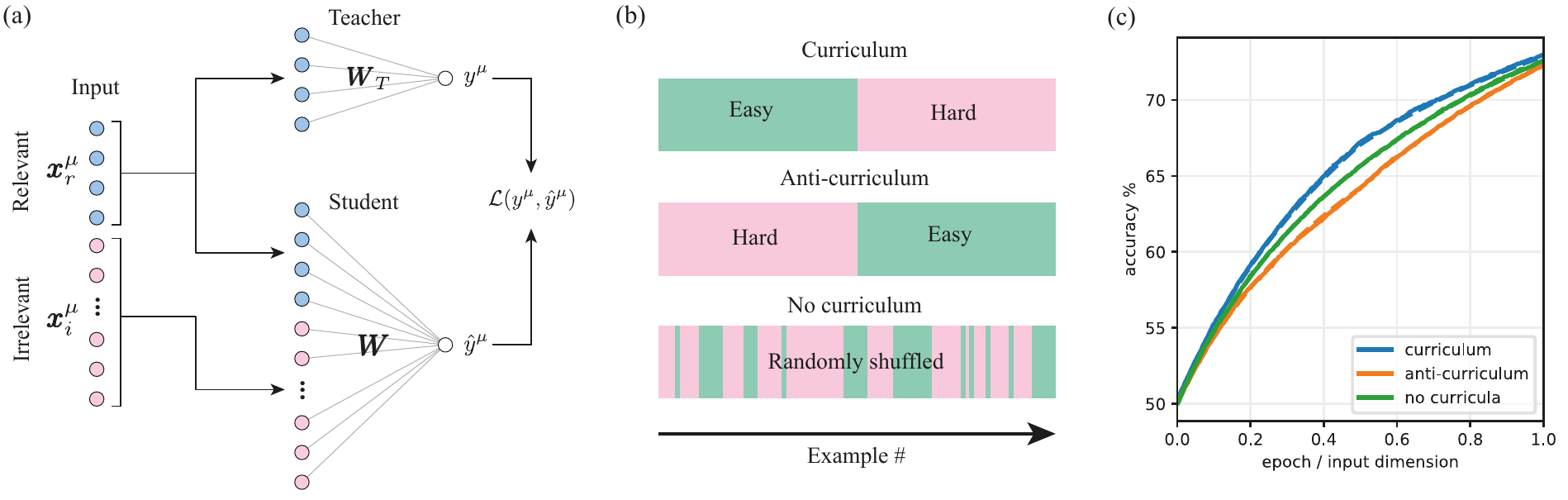}
	\caption{\textbf{Teacher-student setting for curriculum learning.} (a) Illustration of teacher-student setting in which a ``student'' network is trained from \textit{i.i.d.} inputs with labels from a ``teacher'' network. 
	Since the teacher network is sparse, its output depends only on a subset of \emph{relevant} input features.
	(b) We consider curricula which order examples by difficulty, here taken to be the variance in the irrelevant feature dimensions. We refer to increasing, decreasing, and random difficulty order as curriculum, anti-curriculum, and no curriculum, respectively. 
	(c) Example test error on hard examples for the student over training. 
	The switch-point between easy and hard samples lies at $\alpha=1/2$. 
	Solid lines show numerical simulations, while dashed lines show theoretical predictions derived in Section \ref{sec:online_dyn}. For this particular parameter setting, curriculum speeds learning but only modestly improves final performance at $\alpha=1$. Parameters: $\alpha_1=1$, $\alpha_2=1$, $\Delta_1=0$, $\Delta_2=1$, $\gamma=10^{-5}$, $\eta=3$.
	}
	
	\label{fig:approach_overview}
\end{figure}

In the following, we revisit a prototypical model of curriculum learning from \cite{bengio2009curriculum} that finds correspondence to the fading literature \cite{pashler_when_2013} as highlighted in Sec.~\ref{sec:connections}. Our setting is summarised in Fig.~\ref{fig:approach_overview}. The model entails a simple teacher-student setup, where teacher and student are each shallow 1-layer neural networks of size $N$ (also known as perceptrons). The learning task for the student is a binary classification problem, with dataset $\mathcal D = \{(y^\mu,\pmb x^\mu)\}_{\mu=1}^M$, where the ground-truth labels are produced by the teacher network 
$y^\mu = \mathrm{sign}\ \pmb W_T \cdot \pmb x^\mu$. 
The student learns via empirical risk minimisation of an $L_2$ regularised convex loss.

A key feature of this model is that the teacher network is sparse, with only a fraction $\rho<1$ of $\sim\mathcal{N}(0,1)$ non-zero components. Therefore, in order to achieve a good test accuracy, the student has to 
guess which components should be set to zero and align the relevant weights in the correct direction.
A large range of $0<\rho<1$ could give rise to the phenomenology we seek to analyse. In the remainder of the paper we will focus on the case $\rho=0.5$.

We model the variable degree of difficulty in the samples by decomposing each input vector as $\pmb x^\mu = [\pmb x^\mu_{r}, \pmb x^\mu_{i}] \in \mathbb R^N$, where $\pmb x_{r}^\mu\in\mathbb R^{\rho N}$ denotes the relevant components of the input, and $\pmb x_{i}^\mu\in\mathbb R^{(1-\rho) N}$ the irrelevant ones. Note that, crucially, the 
sparse
teacher network is completely blind to the irrelevant part of the input: $y^\mu = \mathrm{sign}\ \sum_{j=1}^{\rho N}W_{T,j} x^\mu_{r,j}$. While $x_{r,j}^\mu$ i.i.d. $\mathcal N(0,1) \,,\forall \mu$,\footnote{In \cite{bengio2009curriculum} the input distribution is uniform between 0 and 1, but this does not qualitatively change the results.} we consider the variance for the irrelevant components to be sample-dependent $x_{i,j}^\mu\sim\mathcal N(0,\Delta^\mu)$. A smaller variance in the irrelevant part induces a higher SNR in the student learning problem. 

The dataset is partitioned according to difficulty levels given by the variances of the irrelevant inputs. For simplicity we consider only two partitions in most of our analysis, but generalisations to multiple difficulty levels follow straightforwardly. We thus have a dataset with $M=(\alpha_1+\alpha_2)N=\alpha N$ samples in total. In the first $\alpha_1 N$ samples the irrelevant inputs have variance $\Delta_1$, while for the remaining $\alpha_2 N$ samples the variance is $\Delta_2>\Delta_1$. In the curriculum learning condition we present the easy examples first, while in the anti-curriculum condition we present the hard examples first. Standard learning presents examples shuffled in random order. 

\section{Online dynamical solution in the large input limit}
\label{sec:online_dyn}

We start by focusing on the same online learning setting explored in \cite{bengio2009curriculum}. We consider a 1-layer student network with sigmoidal activation function, $\sigma(\cdot)=\text{erf}(\cdot/\sqrt2)$, that learns to minimise a mean square error loss with $L_2$ regularisation of intensity $\gamma$, using gradient descent. This yields the updates
\begin{align}
    \label{eq:oSGD_dynamics}
	& \pmb W^{\mu+1}  \! \!\!= \!\! \pmb W^\mu  \!- \! \frac{\eta}{\sqrt{N}} \sigma'\!\left(\!\frac{\pmb W^\mu \! \cdot \! \pmb x^\mu}{\sqrt N}\!\right) \!\left(\!\sigma\!\left(\frac{\pmb W^\mu \! \cdot \! \pmb x^\mu}{\sqrt N}\right) \!- \!y^\mu\!\right) \pmb x^\mu  \!- \! \gamma \pmb W^\mu.
\end{align}

The dynamics of the model can be analysed in the high-dimensional limit $N,M\to\infty$ with $\alpha=M/N=\mathcal{O}(1)$. Generalising the results of \cite{biehl1995learning,saad1995exact} on the online stochastic gradient descent dynamics in single-layer regression problems, we obtain a precise description of the performance at all times, as a function of several order parameters: the squared norm of the relevant and irrelevant part of the student weights $Q_r = \frac1N \pmb W^r \cdot \pmb W^r$  and $Q_i = \frac1N \pmb W^i \cdot \pmb W^i$, respectively; the overlap of the relevant weights of the student and teacher $R = \frac1N \pmb W^r \cdot \pmb W_T$; and the squared norm of the teacher vector $T = \frac1N \pmb W_T \cdot \pmb W_T$. In particular, given $Q_r$, $Q_i$, $R$ and $T$, the test loss (i.e. average loss on a new example) on a dataset with variance $\Delta$ in the irrelevant inputs is given by 
\begin{equation*}
    \begin{split}
	    \mathcal L_\text{MSE} \!&= \!\frac12 \!+ \!\frac1\pi\sin^{-1}\!\!\frac{Q_r+\Delta Q_i}{1+Q_r+\Delta Q_i}
	    \!- \!\frac2\pi\sin^{-1}\!\!\frac{R/\sqrt{T}}{\sqrt{Q_r+\Delta Q_i+1}},
	\end{split}
\end{equation*}
the accuracy by
\begin{equation}
	\mathcal A = \mathbb E\left[ \frac12( y\ \text{sign} \hat y + 1)\right] = \frac12+\frac1\pi\sin^{-1}\left(\frac{R}{\sqrt{T(Q_r+\Delta Q_i)}}\right). \label{eq:accuracy}
\end{equation}
If the dataset contains a random mixture of different difficulty levels $\Delta_1, \Delta_2, \dots$, the loss and accuracy can be obtained by taking a weighted average over the partitions.

To understand how test performance changes through learning, we study the evolution of the order parameters.
Combining their definition with the definition of the dynamics \eqref{eq:oSGD_dynamics} and the fact that the random variables concentrate in the high-dimension as $N\rightarrow\infty$, we obtain an analytic form for the updates: 
$ Q_r \leftarrow  f_{Q_r}\big(Q_r,Q_i,R,T\big),
Q_i \leftarrow  f_{Q_i}\big(Q_r,Q_i,R,T\big),
R \leftarrow  f_{R}\big(Q_r,Q_i,R,T\big); 
$
where $f_{Q_r}$, $f_{Q_i}$ and $f_R$ are long but explicit expressions that are reported in the supplementary material (SM). 

\paragraph*{Dynamical advantages of curriculum.}
With these theoretical results in hand, we can now characterise the performance of curricula in the online setting. We obtain a description of the learning trajectories for each learning protocol, yielding the evolution of training and test accuracies, and of other observables such as the norm of the student and its overlap with the teacher. 

Solving the dynamical equations gives two key advantages relative to simulating models in this setting. First, they are free of finite size effects and stochastic fluctuations. And second, their evaluation is very fast (up to 6 orders of magnitude in simulation time reduction see SM \ref{sm:speedup}), enabling systematic exploration of the parameter space of the problem, along with fine-grained optimisation over hyper-parameters such as learning rate, weight decay and scaling in the initialisation. 

Optimising final test accuracy separately for each curriculum strategy, we find that curriculum learning is the optimal strategy, followed by baseline (no-curriculum) and lastly anti-curriculum. In Fig.~\ref{fig:approach_overview}c we show typical learning trajectories for a dataset with equal numbers of easy and hard samples. The results of the simulations (solid lines) are well-described by our theoretical equations (dashed lines), and show that the curriculum strategy leads to better performance throughout training. Fig.~\ref{fig:approach_overview}c shows the evolution during training of the test accuracy computed on the whole dataset.

\begin{figure}
	\centering
	\begin{subfigure}[b]{.30\linewidth}
        \centering
        \includegraphics[width=\linewidth]{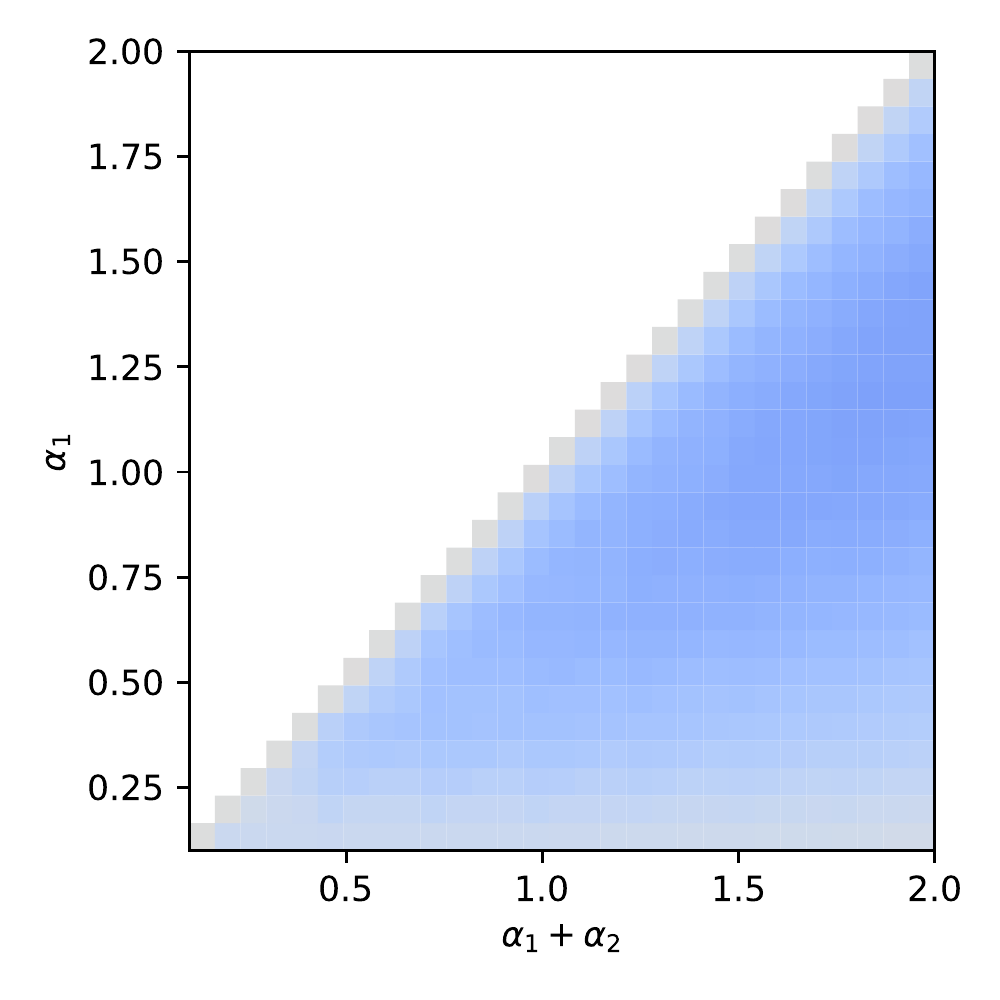}
		\caption{Curriculum learning.}
    \end{subfigure}
    \centering
	\begin{subfigure}[b]{.30\linewidth}
        \centering
        \includegraphics[width=\linewidth]{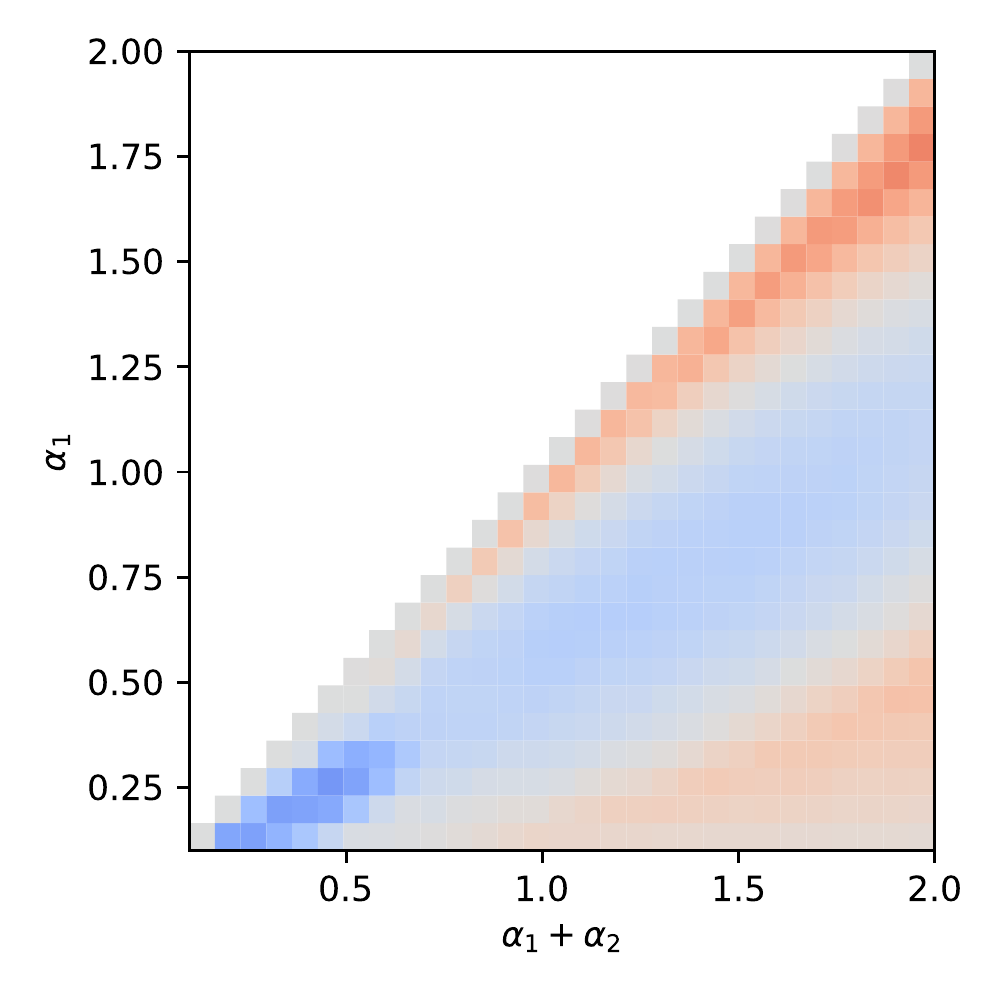}
		\caption{Anti-curriculum learning.}
    \end{subfigure}
	\begin{subfigure}[b]{.37\linewidth}
        \centering
        \includegraphics[width=\linewidth]{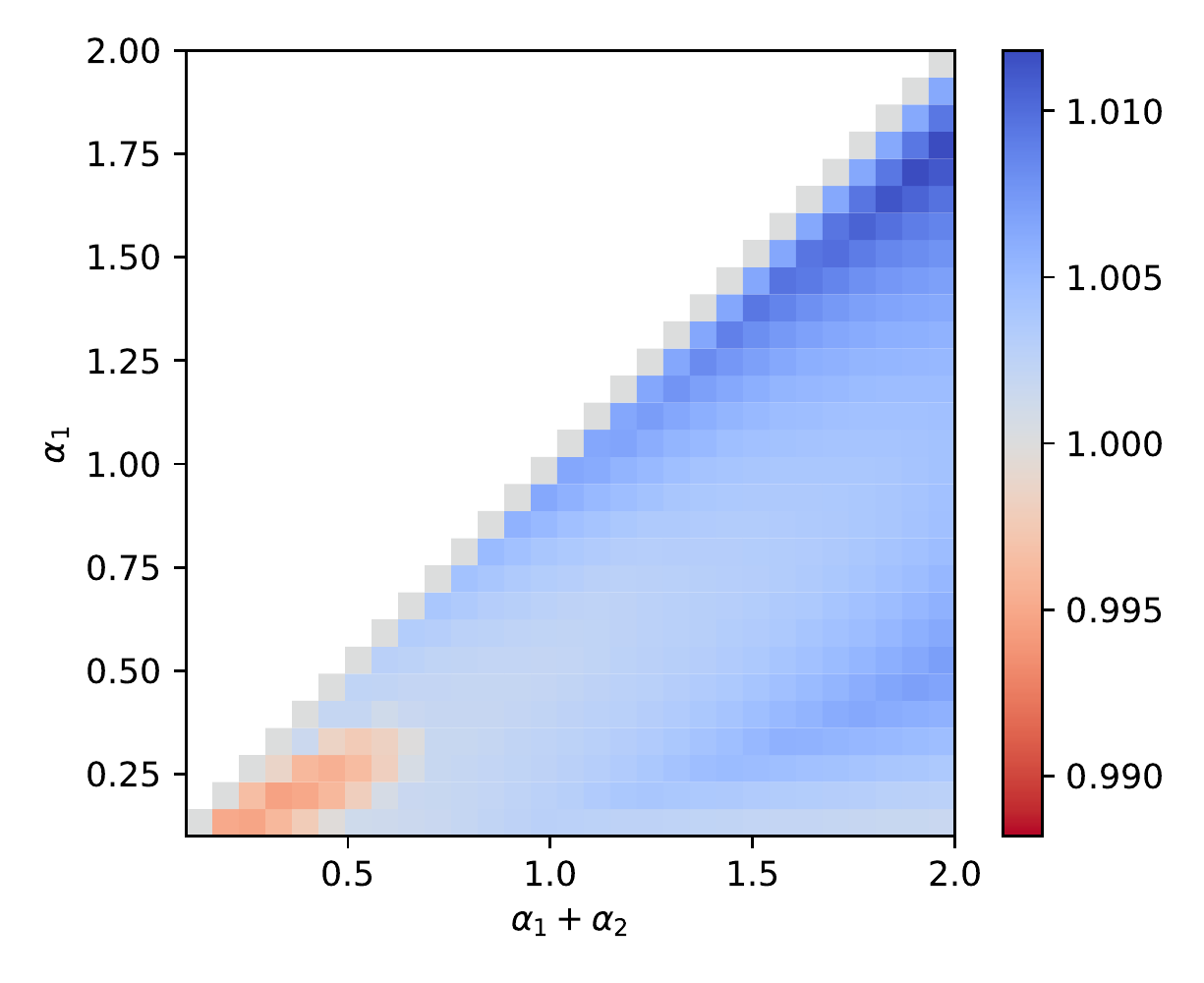}
		\caption{Curriculum vs anti-curriculum.}
    \end{subfigure}
	\caption{\textbf{Phase diagram of online learning performance gap with optimal parameters.} The colour scale shows the ratio of the accuracy on hard instances reached by curriculum over no-curriculum (a), anti-curriculum over no-curriculum (b), and curriculum over anti-curriculum (c), as a function of the total dataset size ($\alpha_1 + \alpha_2$) and easy dataset size ($\alpha_1$). Curriculum broadly benefits performance and anti-curriculum is effective in certain regions, but the size of the improvement is modest. Parameters: $\rho=0.50,\Delta_1=0,\Delta_2=1$.
	}
	\label{fig:classification_curricula_comparison_heatmap}
\end{figure}

Next, we systematically trace the effect of curriculum for a range of total dataset sizes ($\alpha_1+\alpha_2$) and number of easy examples $\alpha_1$ in the phase diagram in Fig.~\ref{fig:classification_curricula_comparison_heatmap}. This diagram shows in panels (a) and (b) the accuracies on hard instances reached at the end of training, by curriculum learning and anti-curriculum learning respectively, normalised by the accuracy reached by the standard strategy. The two heatmaps show that curriculum learning always outperforms standard learning and that, on the other hand, anti-curriculum learning outperforms standard learning only in part of the diagram. 
Comparing the two strategies, in Fig.~\ref{fig:classification_curricula_comparison_heatmap} (c), we can observe that there is a region for small $\alpha$ and $\alpha_1$ where anti-curriculum learning is the best strategy, while in the majority of the situations curriculum learning is best. Interestingly, there is a sizeable region of the diagram in which \textit{both} curriculum and anti-curriculum help, possibly explaining why both have been recommended in prior work \cite{bengio2009curriculum,HacohenW19,kocmi_curriculum_2017,zhang2018empirical,zhang_curriculum_2019}. A possible intuition behind this counter-intuitive phenomenon highlighted by our analysis is that, in some settings, the large amount of noise contained in the hard data will always be too disruptive for effective learning. Thus, leaving the easy (cleaner) data for last could allow the model to better exploit it.

Further, we find that our setting, in which a small task-relevant signal is embedded in large task-irrelevant variation, is critical to the benefit of curriculum. Fig.~\ref{fig:coupling_effect} shows performance as a function of sparsity $\rho$, additional details are deferred in the SM~\ref{si:sparsity}. Non-sparse tasks do not benefit. Hence curriculum aids tasks with many irrelevant factors of variation. Interestingly, the literature from human psychology shows precisely this: no curriculum benefits for low-dimensional tasks or tasks with no variation in irrelevant dimensions \cite{pashler_when_2013}. 

\begin{figure}
	\centering
	\begin{subfigure}[b]{.304\linewidth}
        \centering
        \includegraphics[width=\linewidth]{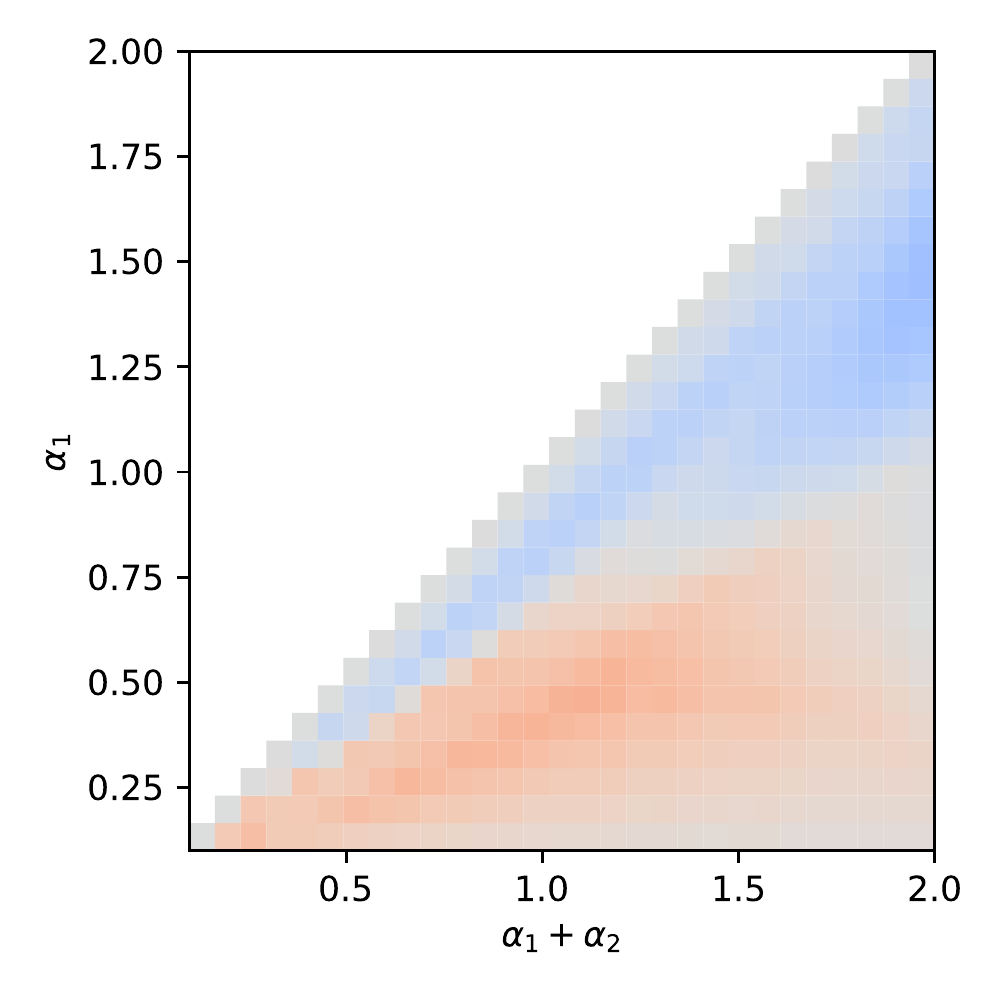}
		\caption{Curriculum learning.}
    \end{subfigure}
    \centering
	\begin{subfigure}[b]{.37\linewidth}
        \centering
        \includegraphics[width=\linewidth]{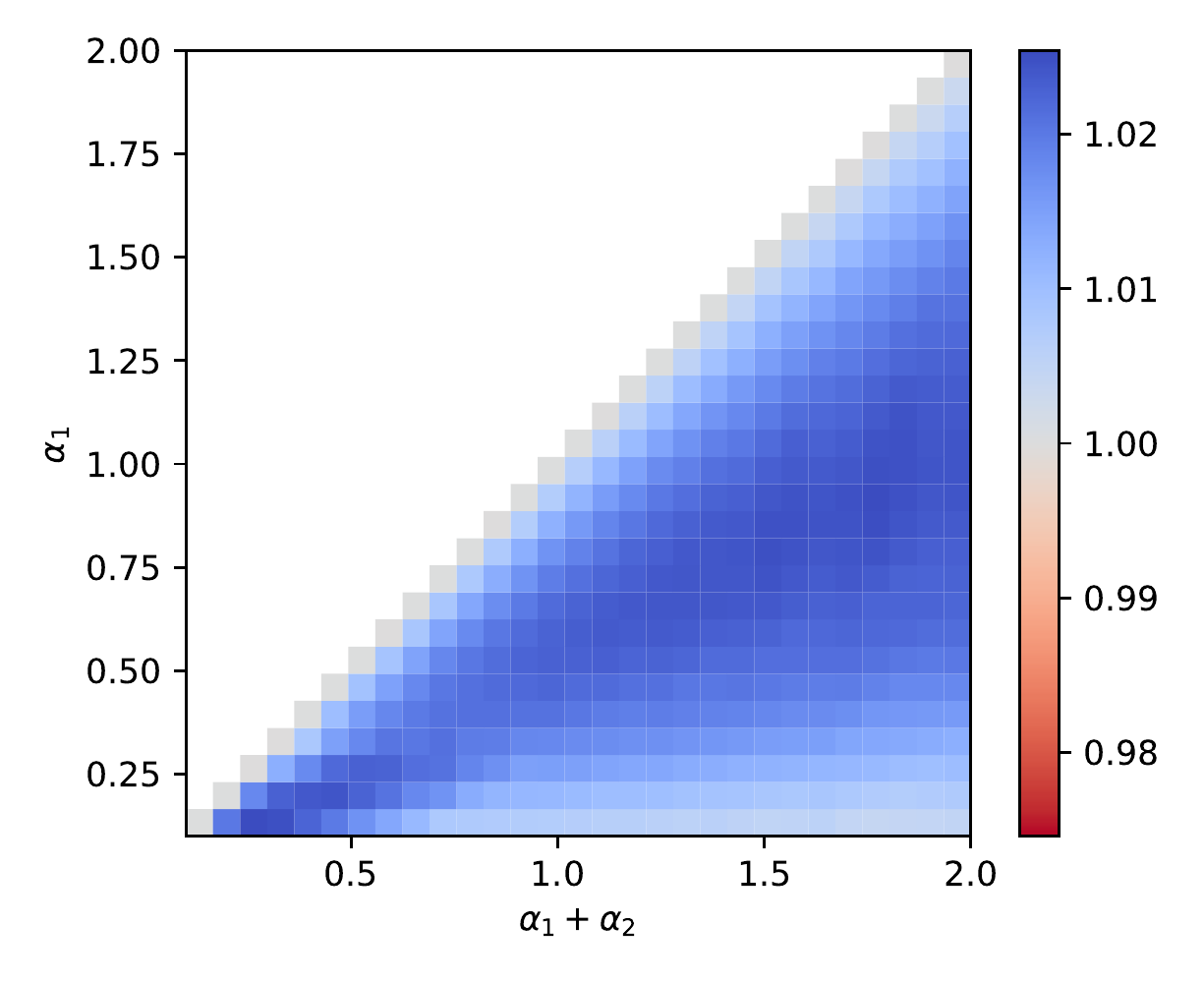}
		\caption{Anti-curriculum learning.}
    \end{subfigure}
	\centering
	\begin{subfigure}[b]{.304\linewidth}
        \centering
        \includegraphics[width=\linewidth]{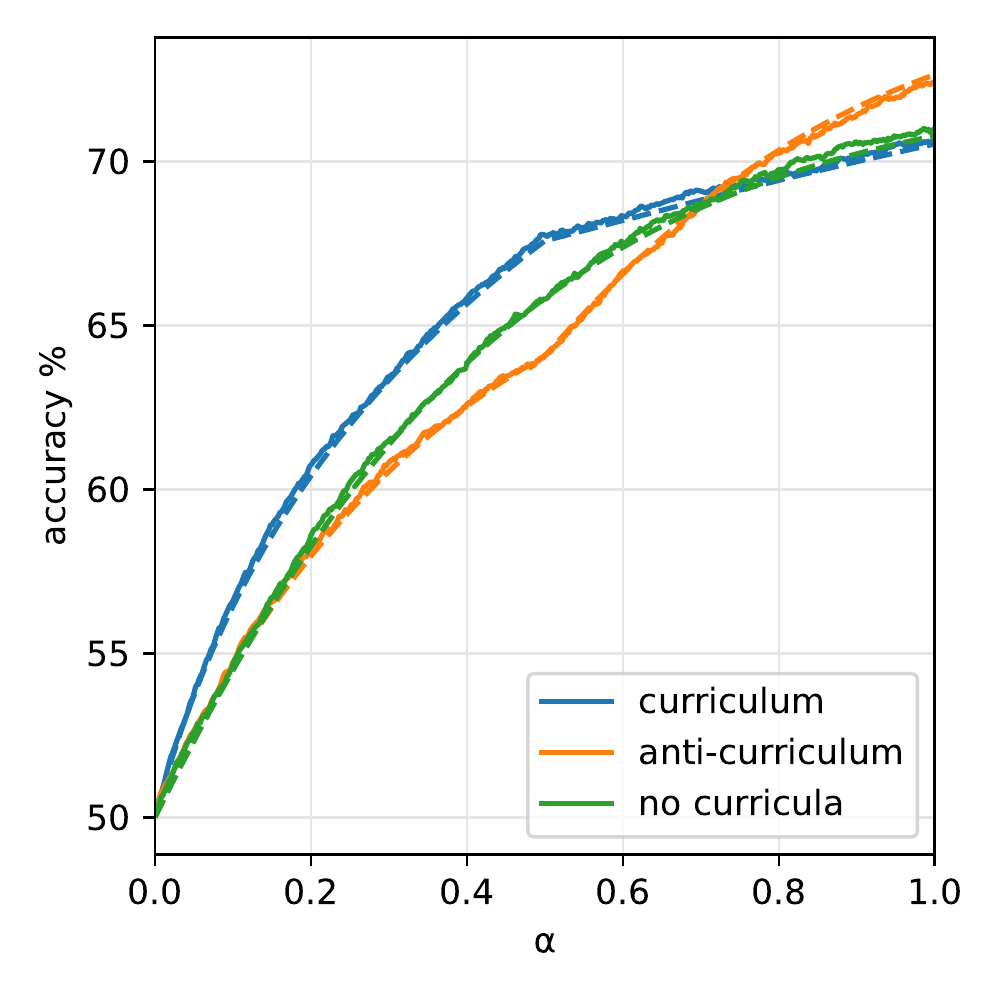}
		\caption{Accuracy trajectories.}
    \end{subfigure}
	\caption{\textbf{Performance gap starting from high initialisation norm.} The first two figures show the accuracy-gap on hard instances between curriculum learning and the baseline (a) and anti-curriculum learning and the baseline (b). 
	Contrary to the phase diagram in Fig.~\ref{fig:classification_curricula_comparison_heatmap}, curriculum learning is not always the optimal and anti-curriculum is not always the worst strategy. 
	The right panel shows the accuracy evaluated on the hard samples for $\alpha_1=\alpha_2=0.5$. 
	}
	\label{fig:classification_curricula_comparison_heatmap_highinit}
\end{figure}

Our results also highlight the intricate dependence of curriculum on parameters of the learning setup. If not all parameters are correctly optimised, we can observe more complex scenarios. For instance, the initialisation condition for the norm of the weights of the student plays an important role. We explore this dependence by changing the variance of the normal distribution from which the initial weights are sampled from. We observe that anti-curriculum learning becomes the best strategy when the variance is large, as shown in Fig.~\ref{fig:classification_curricula_comparison_heatmap_highinit} for weights of order 1. In this case, curriculum learning shows an advantage only in the first phase when easy examples are shown, which is consistent with the results of \cite{weinshall2020theory}. However, in the next phase when hard examples are shown, the curriculum strategy does not extract enough information and it is outperformed by the other two strategies. The fact that curriculum or anti-curriculum can look better depending on the parameter setting might help explain the confusion in the literature over the best protocol \cite{bengio2009curriculum,HacohenW19,kocmi_curriculum_2017,zhang2018empirical,zhang_curriculum_2019}.
At least in this model, better performance from anti-curriculum is a signature of a sub-optimal choice of the parameters. 

To summarise our findings in this online learning setting, curriculum mainly offers a \textit{dynamical advantage}: it speeds up learning but has minimal impact on asymptotic performance. 

\section{Batch learning solution}\label{sec:batch}

The previous section discussed the online case where each example is used once and then discarded. However, in common machine learning practice, neural networks typically revisit each sample repeatedly until convergence. Therefore an important question is: \textit{can curricula lead to a generalisation improvement when trained on the same dataset until convergence?}

We investigate this question by considering a student that learns from slices of a dataset in distinct optimisation phases, where in each phase the student optimises a $L_2$-regularised logistic loss. Without further modification, curriculum can have no effect in this setting: due to the convex nature of the teacher-student setup \cite{engel2001statistical}, the network is bound to converge to a minimum uniquely determined by the final slice of data, with no memory of the progress made at intermediate steps. 
This simple observation may help explain empirical observations on real data, such as \cite{wu2020curricula}, which find no benefit of curriculum in standard settings. In fact, in principle curriculum could still influence non-convex problems \cite{bengio2009curriculum} but empirical results in the ML field are not showing clear signals of memory retention. A possible explanation of this is that relying on dynamical memory effects requires careful tuning of the learning rate and of the number of training epochs, while typical choices for these hyper-parameters could lead to memory loss and performance inconsistencies.
These observations raise the theoretical question of how to better implement curriculum learning to induce a non-vanishing effect also in batch learning settings.

To instantiate a long-term memory effect in our model, we propose biasing the optimisation landscape via a Gaussian prior, centred around the optimiser of the previous learning phase. The additional term in the loss acts as an elastic coupling between the successive phases, and the associated intensity $\gamma_{12}$ is then an additional hyper-parameter of the model. This scheme is similar to regularisation methods proposed against catastrophic interference in continual learning, such as Synaptic Intelligence \cite{pmlr-v70-zenke17a}. Changing the loss according to the curriculum prescription effectively makes the learning algorithm \textit{aware} of the different levels of difficulty in the dataset.

Tools from statistical physics can be used to analytically compute test performance under this scheme. In order to simplify the presentation, we first consider just two learning phases. It is  natural to frame this setting as a 2-level problem, involving two systems with independent copies of the network weights $\pmb{W}_1$ and $\pmb{W}_2$. In a typical statistical physics approach, we associate a Boltzmann-Gibbs measure to the systems, with an energy function determined by the regularised logistic loss $\mathcal{L}_\gamma$. While the statistical properties of the first system can be determined self-consistently, the added elastic interaction creates a dependence of the second measure on the configurations of the first system. In mathematical terms, the coupled system is represented by the following partition function:
\begin{equation}
    \begin{split}
        \langle Z & (\pmb{W}_2, \pmb{W}_1; \mathcal{D}_1, \mathcal{D}_2) \rangle_{\pmb{W}_1} \!=\! \int d\pmb{W}_1 \frac{e^{-\beta_1 \mathcal{L}_{\gamma_1}(\pmb{W}_1, \mathcal{D}_1)}}{Z_1(\pmb{W}_1)} 
        \log \int d\pmb{W}_2 \,e^{-\beta_2 \left( \mathcal{L}_{\gamma_2}(\pmb{W}_2,\mathcal{D}_2) + \frac{\gamma_{12}}{2} \lVert \pmb{W}_2 - \pmb{W}_1 \rVert^2_2 \right)}
    \end{split}
\end{equation}
where $\mathcal{D}_1,\mathcal{D}_2$ denote the two dataset slices. This object represents the normalisation of the Boltzmann-Gibbs measure, and allows one to extract relevant information on the asymptotic behaviour of our model. The optimisations entailed in each learning phase can be described in the ``low noise'' limit of $\beta_1,\beta_2\to\infty$, where the measures focus on the minimisers of the respective losses. In order to study a self-averaging quantity that does not depend on a specific realisation of the dataset, we aim to compute the associated average free-energy:
\begin{equation}
    \Phi = \lim_{N\to\infty}\lim_{\beta_1,\beta_2\to\infty} \frac{1}{\beta_2 N} \left< \log \left< Z(\pmb{W}_2, \pmb{W}_1; \mathcal{D}_1, \mathcal{D}_2) \right>_{\pmb{W}_1} \right>_{\mathcal{D}_1, \mathcal{D}_2}.
\end{equation}
This quantity can be seen as a special case of the so-called Franz-Parisi potential computation \cite{franz1997phase, saglietti2020solvable}, and the entailed double average can be evaluated through the replica method. Refer to SM for details. 

\begin{figure}
    \centering
	\begin{subfigure}[b]{.4\linewidth}
        \centering
        \includegraphics[width=\linewidth]{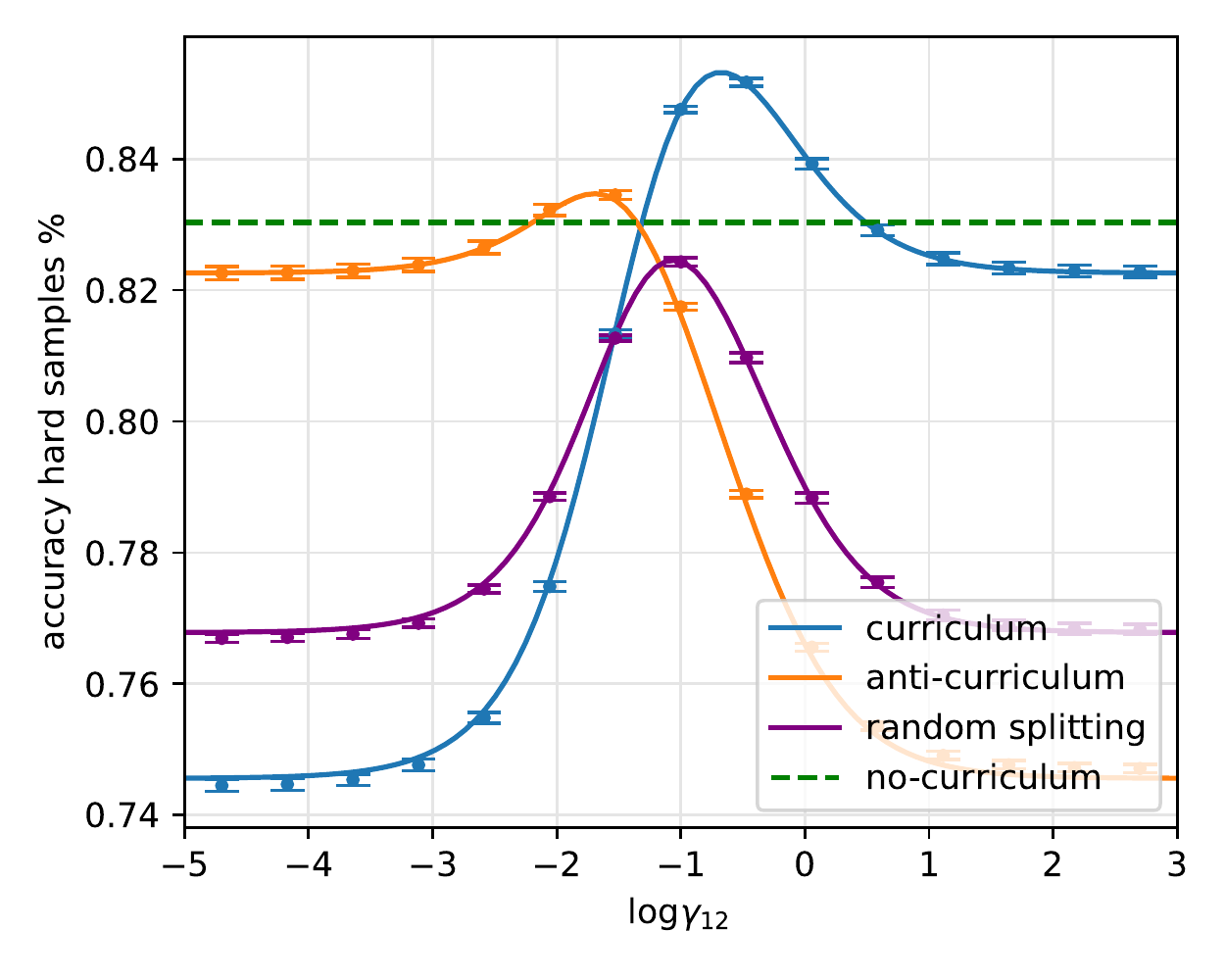}
		\caption{Learning with curricula}
    \end{subfigure}
	\centering
	\begin{subfigure}[b]{.4\linewidth}
        \centering
        \includegraphics[width=\linewidth]{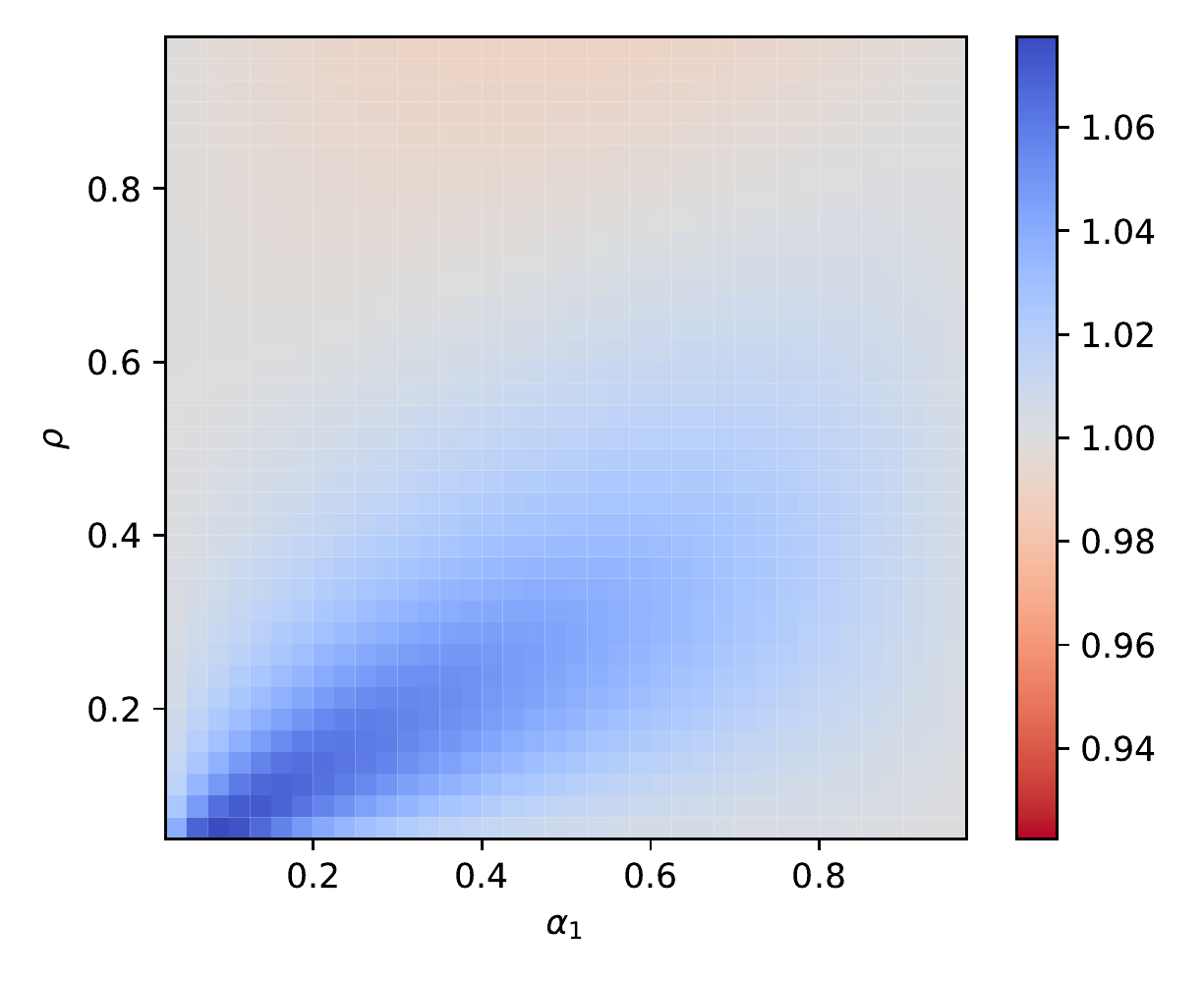}
		\caption{Curriculum vs no-curriculum}
    \end{subfigure}
	\caption{
	\textbf{Effect of elastic coupling (Gaussian prior) between curriculum phases.}  (a) comparison between asymptotic performance of curricula (full lines) and single batch learning, at $\alpha_1=1$ $\alpha_2=1$, with a regularisation $\gamma_1$ that yields the best generalisation when learning the entire dataset (in principle not optimal for the other strategies). The points represent the results from $10$ numerical simulations at size $N=2000$. Parameters: $\rho=0.50$, $\Delta_1=0$ and $\Delta_2=1$.
    (b) ratio between the accuracy reached by curriculum learning over anti-curriculum as a function of the number of easy samples in a dataset of dimension $\alpha_1+\alpha_2=1$, and of the sparsity level of the teacher $\rho$. Note that $\rho$ can also be seen as the fraction of relevant components in the inputs. $\Delta_1=0$ and $\Delta_1=1$. $\gamma_1=\gamma_2$ and $\gamma_{12}$ where set the values that optimise test performance.
	}
	\label{fig:coupling_effect}
\end{figure}

Similar to the online case, in high-dimensions the free-entropy concentrates on a deterministic function that depends on several order parameters that capture the geometrical distribution of teacher and student configurations. In addition to those already introduced in Sec.~\ref{sec:online_dyn}, we also have $\delta Q$, which is linked to the variance of the student norm. Moreover, for each order parameter we also need to introduce a conjugate parameter, denoted in the following with the hat symbol. The final expression for the free-energy reads:  
\begin{equation}
    \begin{split}
        \Phi =\mathrm{extr} & \Big[-\left(\hat{R}R+\frac{1}{2}\left(\left(\hat{Q}\delta Q-\delta\hat{Q}Q\right)_{r+i}\right)\right) 
        +\,g_{S}(\gamma_1,\gamma_2,\gamma_{12})+\alpha_{1}\,g_{E}\left(\Delta_1\right)+\alpha_{2}\,g_{E}\left(\Delta_2\right)\Big]
    \end{split}
\end{equation}
where $g_S$ and $g_E$ are two scalar functions, often called entropic and energetic channels, that encode the dependence of the optimisation problem on the Gaussian prior and the logistic loss respectively.
The extremum condition for the free-energy yields a system of fixed-point equations that converge to an asymptotic prediction for the order parameters, comparable with the results of numerical simulations on large instances, Fig.~\ref{fig:coupling_effect}. At convergence, the order parameters can be inserted again in Eq.~\ref{eq:accuracy} to obtain an estimate of the test accuracy. Note that this formalism is not limited to two phases, but can be extended to the case of a discrete number of sequential stages. 

\paragraph*{The importance of sparsity.}\label{sec:sparsity}
Sparsity is a key ingredient in determining the impact of curriculum strategies. It naturally introduces a notion of relevant and irrelevant inputs, and defines a secondary learning goal: identifying what part of the presented data should be disregarded by the model.
Curriculum learning can aid this identification process, since the easy samples are more transparent to this structure. This is also observed in human experiments \cite{pashler_when_2013}. 
However, the relative difficulty of the problem of inferring the support of the teacher and the problem of aligning with its non-zero components depends on the degree of sparsity $\rho$, so the effectiveness of curriculum can vary with it. 

In the right panel of Fig.~\ref{fig:coupling_effect}, we explore the interplay between the sparsity of the teacher $\rho$ and the fraction of easy samples in the dataset $\alpha_1$, comparing curriculum with the no-curriculum baseline. The phase diagram highlights the variability in the impact of the curriculum ordering: 
\begin{itemize}
    \item Curriculum is most effective at low values of $\rho$ and close to the diagonal, where the fraction of easy examples in the dataset is comparable to the fraction of relevant dimensions. 
    \item When $\rho>0.5$, the possible gain from ordering the samples according to difficulty is counterbalanced by the instrinsic cost of splitting 
    the information content into two blocks,
    thus curriculum can become detrimental.
    \item When $\alpha_1$ is too small compared to $\rho$ (above diagonal), the first stage in the curriculum strategy can only help in the support identification problem, but will not allow a good estimation of the direction of the teacher. Because of the elastic prior, the second stage cannot improve too much over it and the effect of curriculum is small.
    \item When $\alpha$ is larger than the sparsity (below diagonal), the easy examples contain sufficient information for solving both the support and the teacher estimation problems, and this information is also exploited by the baseline. Thus the improvement of curriculum becomes negligible. 
\end{itemize}
We refer to the SM for an in-depth comparison with anti-curriculum.

\begin{figure}[h]
	\centering
	\begin{subfigure}[b]{.30\linewidth}
        \centering
        \includegraphics[width=\linewidth]{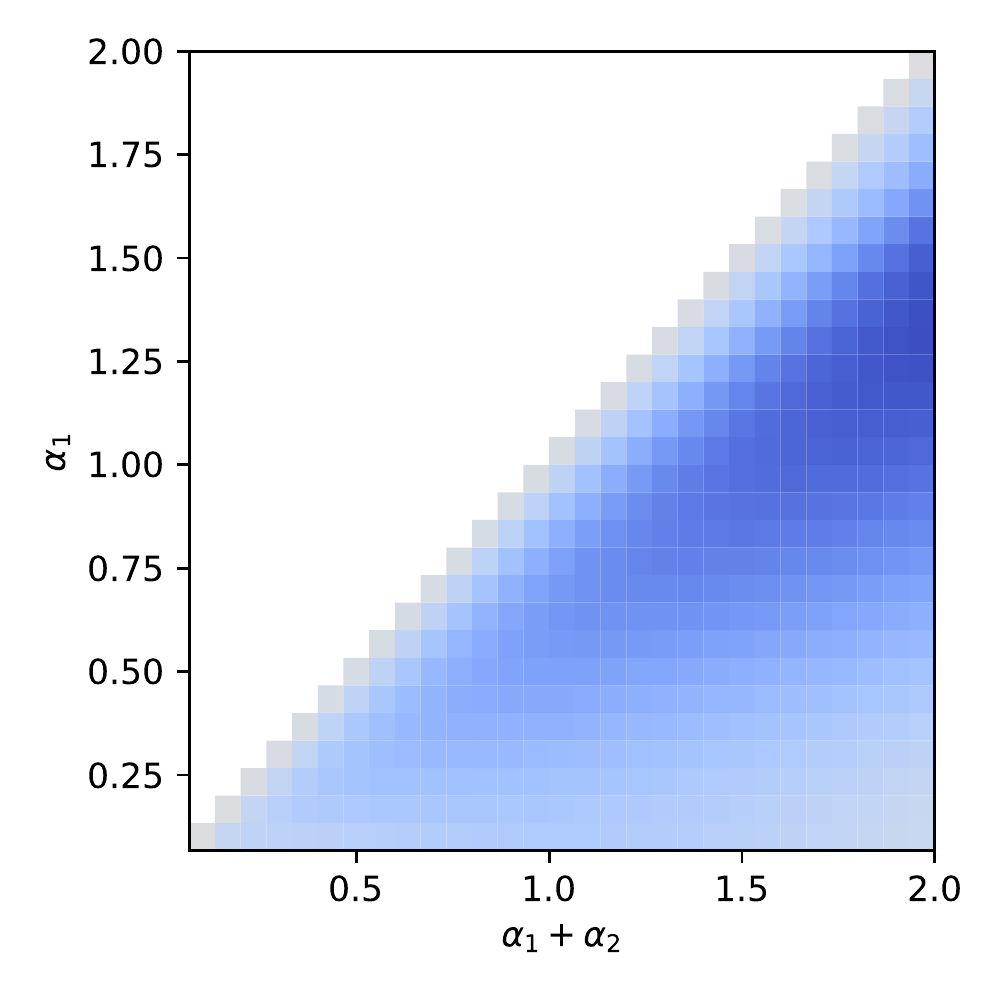}
		\caption{Curriculum learning.}
    \end{subfigure}
    \centering
	\begin{subfigure}[b]{.30\linewidth}
        \centering
        \includegraphics[width=\linewidth]{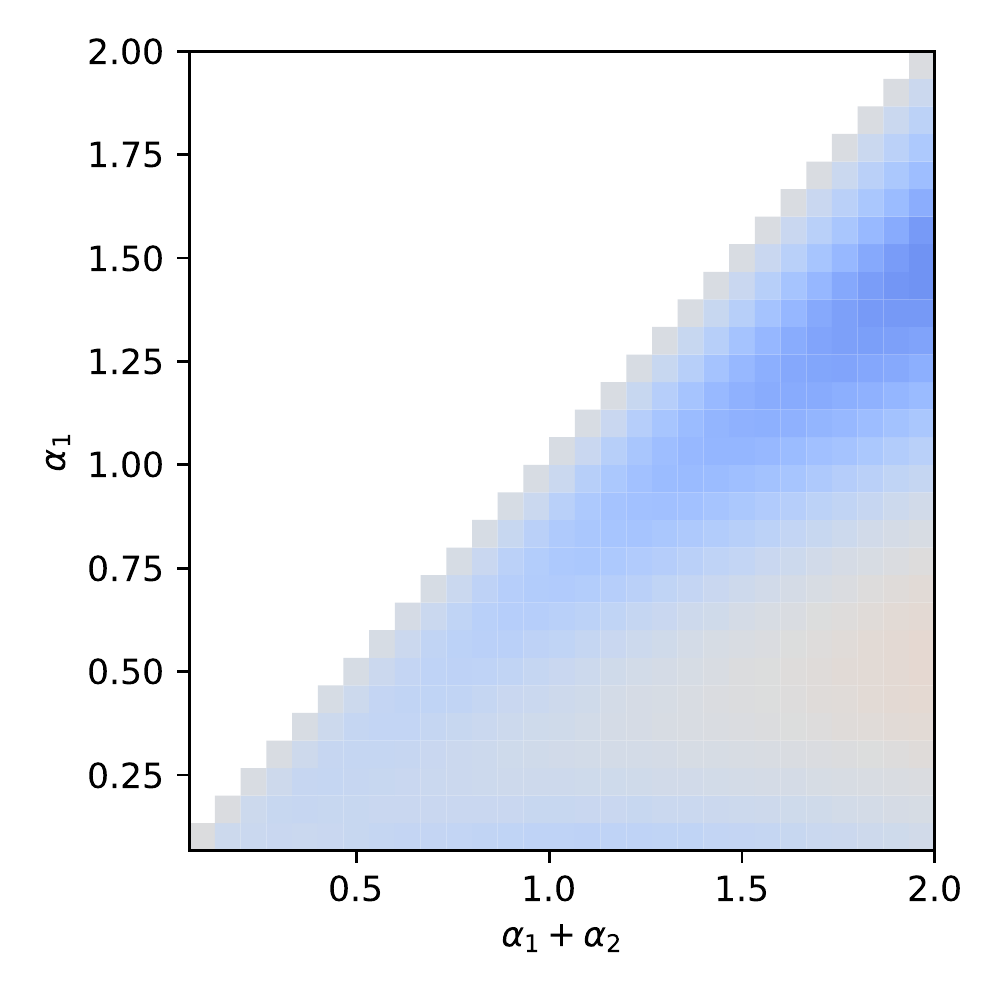}
		\caption{Anti-curriculum learning.}
    \end{subfigure}
	\begin{subfigure}[b]{.37\linewidth}
        \centering
        \includegraphics[width=\linewidth]{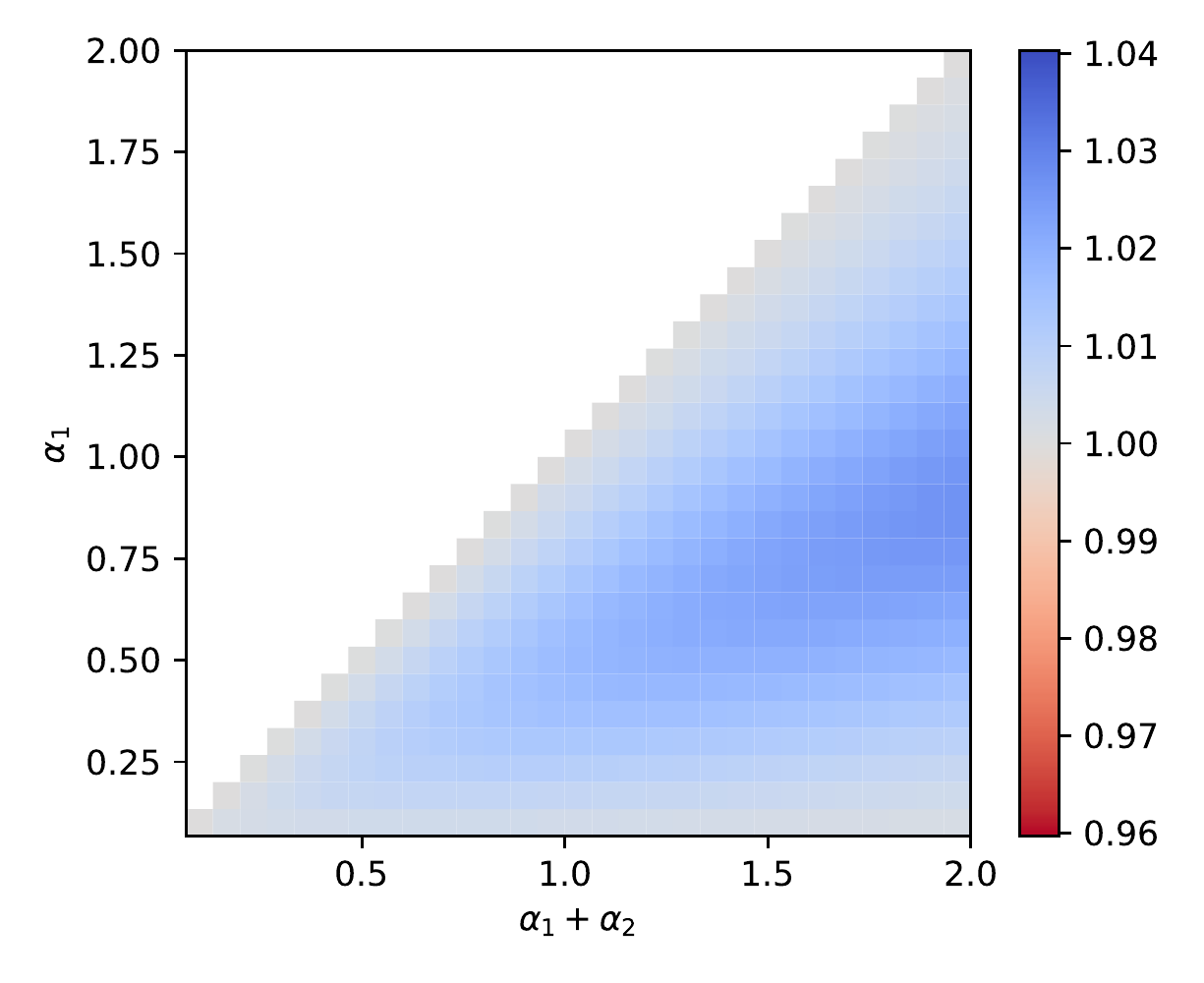}
		\caption{Curriculum vs anti-curriculum.}
    \end{subfigure}
    \caption{\textbf{Phase diagram for the performance gap in the batch setting.} The colour scale shows the ratio of the accuracy on hard instances for curriculum over no-curriculum (a), anti-curriculum over no-curriculum (b), and curriculum over anti-curriculum (c), as a function of the total dataset size ($\alpha_1 + \alpha_2$) and easy dataset size ($\alpha_1$). In contrast to the online case, performance benefits are greater and curriculum is strictly better than anti-curriculum. Both $\gamma_1=\gamma_2$ and $\gamma_{12}$ are optimised point-wise, in order to yield the best test accuracy. Parameters: $\rho=0.50,\Delta_1=0,\Delta_2=1$.
    }
	\label{fig:all_optimized_2_stages}
\end{figure}

\paragraph*{Asymptotic advantages of curriculum.}

Contrary to the case of online SGD, if the fraction of relevant directions is small, batch learning with elastic coupling notably improves test accuracy of both curriculum and anti-curriculum above the baseline. This confirms the utility of curriculum strategies when the signal is partially "hidden in clutter" \cite{clerkin_real-world_2017}. 

Fig.~\ref{fig:all_optimized_2_stages} shows similar phase diagrams to Fig.~\ref{fig:classification_curricula_comparison_heatmap} but for the batch setting. At each point in the phase diagram the regularisation level $\gamma_1=\gamma_2$ and the coupling $\gamma_{12}$ are optimised to yield the best accuracy.
We find that the performance order is nearly always preserved: curriculum followed by anti-curriculum followed by baseline. In the SM we see similar improvements by applying the elastic coupling strategy both in the online setting and on real data.

In summary, in the batch setting, splitting the learning process in stages might not be advantageous per se. However, our observations show that if the loss is modified to reduce memory loss between the learning stages, curriculum learning strategies can offer a measurable \textit{asymptotic advantage}.

\section{Connection with experimental literature}\label{sec:connections}

\begin{figure}
	\centering
	\begin{subfigure}[b]{.35\linewidth}
        \centering
        \includegraphics[width=\linewidth,trim={0.8cm 0.8cm 0.5cm 0.5cm},clip]{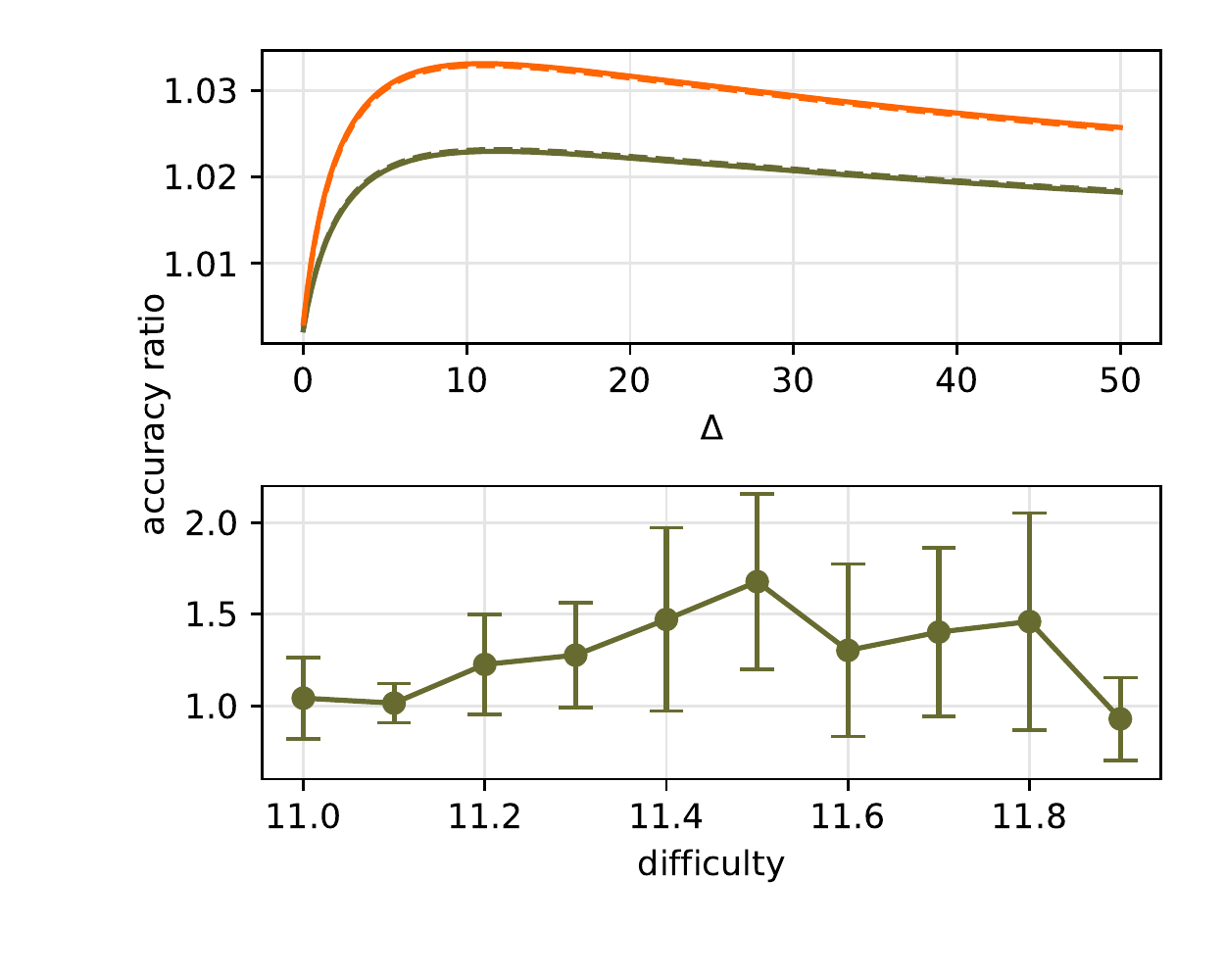}
		\caption{Generalisation gap per difficulty.}
    \end{subfigure}
	\centering
	\begin{subfigure}[b]{.35\linewidth}
        \centering
        \includegraphics[width=\linewidth,trim={0.8cm 0.8cm 0.5cm 0.5cm},clip]{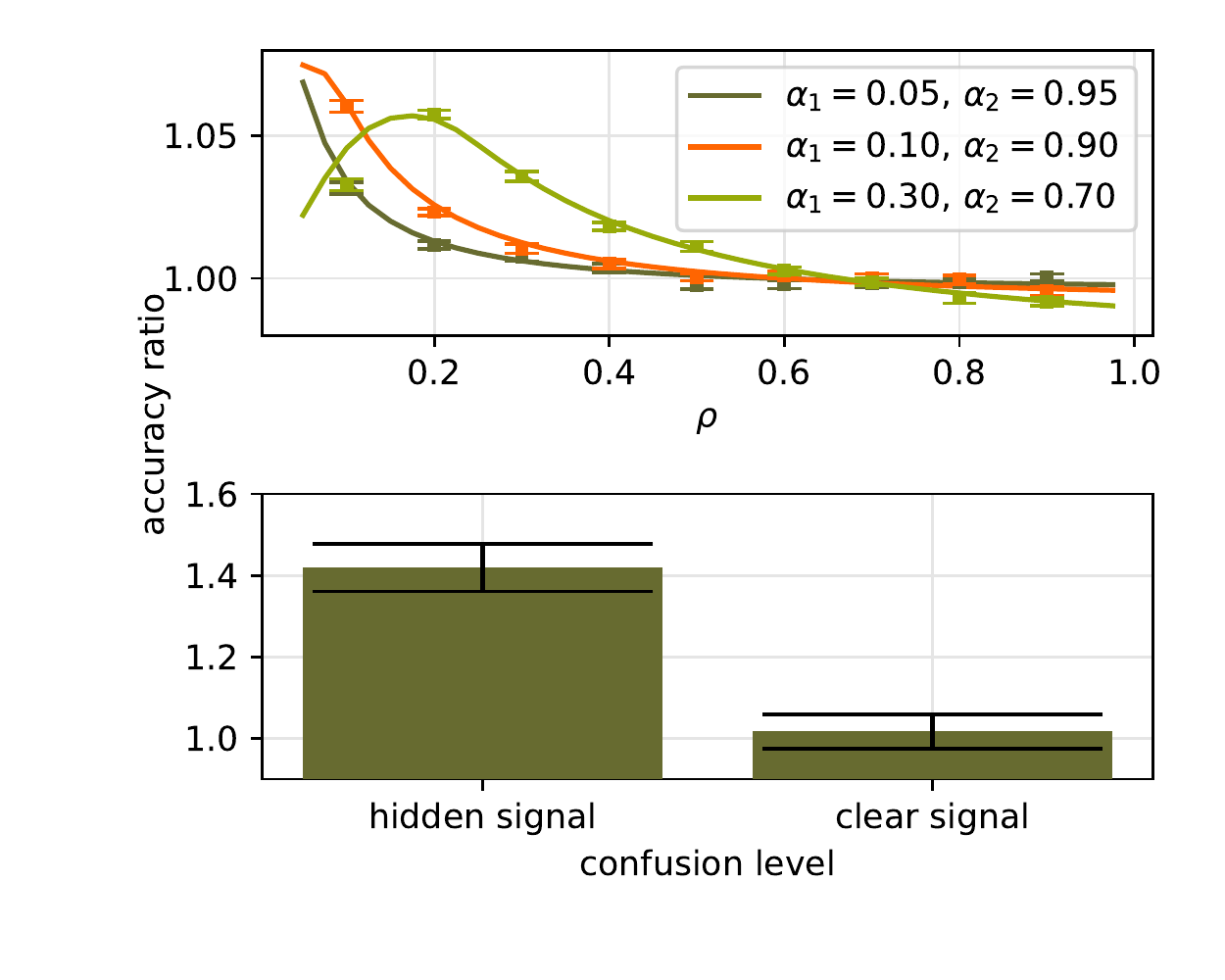}
		\caption{Role of sparsity.}
    \end{subfigure}
	\centering
	\begin{subfigure}[b]{.28\linewidth}
        \centering
        \includegraphics[width=\linewidth,trim={0cm -1cm 0cm 0cm},clip]{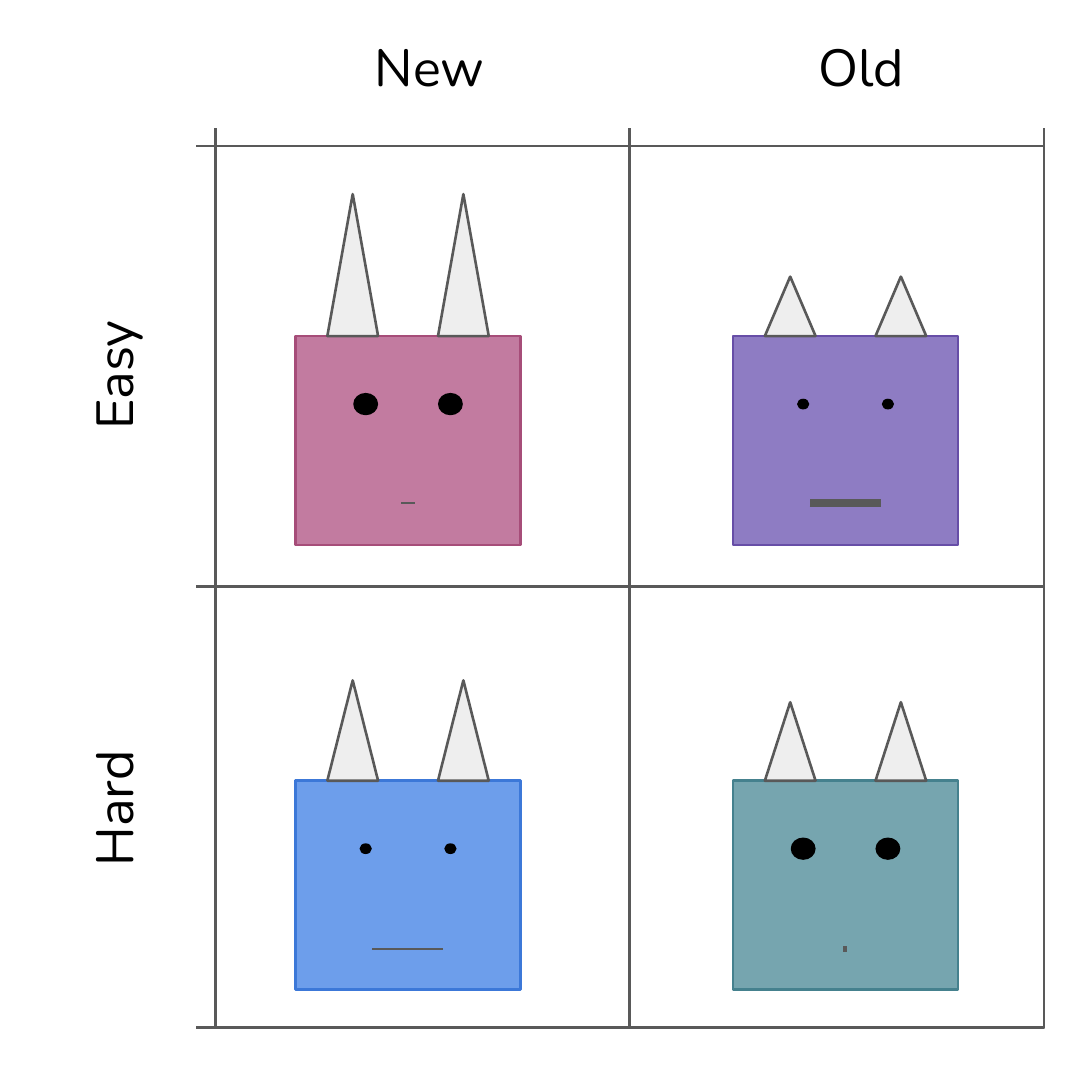}
		\caption{Fading experiment \cite{pashler_when_2013}.}
    \end{subfigure}
	\caption{\textbf{Connection with psychology experiments.}
	(a) Top: Accuracy ratio of different strategies in the model, with curriculum/no-curriculum in green and curriculum/anti-curriculum in orange. The ratio shows non-monotonic behaviour. Bottom: The accuracy ratio obtained by \cite{liu2008easy}.
	Parameters $\rho=0.5$, $\Delta_1=0.0$, $\Delta_2=1.0$, $\alpha_1=1$, $\alpha_2=1$ and optimal learning rate, norm at initialisation and weight decay intensity.
	(b) Top: Dependence on the sparsity of the generalisation gain of curriculum over no-curriculum, measured as ratio between final accuracy, for fixed total dataset size ($\alpha_1+\alpha_2=1$). Bottom: The ratio obtained from experiments 3 and 4 of \cite{pashler_when_2013}.
	(c)
	Example cartoon stimuli from the ``fading'' paradigm used in \cite{pashler_when_2013}, where participants distinguish daemons of the old world from daemons of the new world. The distinguishing feature (horn length) is diluted among many irrelevant features (colour, eye size, mouth size). Highlighting the relevant feature to participants leads to better and faster learning. 
	}
	\vspace{-1mm}
	\label{fig:psychology}
\end{figure}

Recent work has suggested that curriculum learning could provide an important window into the learning algorithms at work in biology \cite{kepple2022}. Our analysis makes several predictions for curriculum effects. In this section we assess these predictions based on connections to extant experiments and propose future experimental tests. 

First, we find that a curriculum strategy yields a speed up in learning in all the tested settings (see Fig.~\ref{fig:approach_overview}c). This acceleration is broadly consistent with the findings from cognitive science \cite{lawrence1952transfer,baker1954discrimination,pashler_when_2013}.
By contrast, our results show that the speed improvement does not necessarily translate into a sizeable generalisation error improvement, and the performance achieved at the end of training can even deteriorate when learning hyperparameters are not fully optimised (c.f. Fig.~\ref{fig:classification_curricula_comparison_heatmap_highinit}). Deterioration due to curricula has generally not been reported in the psychology literature, though it has been observed in ML \cite{wu2020curricula}. This fact may suggest that animals naturally learn with near-optimal hyperparameters such that curricula generally confer benefits. 

A more specific observation concerns the performance on different difficulties after learning. As reported in \cite{liu2008easy}, human and rodent subjects trained in an auditory task using curricula showed the greatest improvement for intermediate levels of difficulty as depicted in Fig.~\ref{fig:psychology}a bottom panel. The same conclusion can be drawn from the experiment of \cite{hornsby_improved_2014,roads_easy--hard_2018}, where, surprisingly, subjects trained with curricula to classify medical images showed poor performance in hard tasks compared to the control group. To address this phenomenon, we calculate accuracy as a function of difficulty in the model in Fig.~\ref{fig:psychology}a top panel. Consistent with these experiments, we find regimes where the gap between curriculum learning and the baseline is non-monotonic, with the largest performance gain for intermediate difficulties. Contrary to \cite{hornsby_improved_2014,roads_easy--hard_2018}, however, we do not observe negative effects of curriculum for high difficulties. Further experiments that more systematically manipulate training and transfer difficulties could provide a stronger test of these predictions.

A key ingredient in our model is the role of sparsity, such that a small signal is embedded amidst many irrelevant features. Experimentally, the importance of having many factors of variation to obtaining a curriculum effect has been documented in the ``fading'' experiments of \cite{pashler_when_2013}. Human subjects were trained on classification tasks involving stimuli with one task-relevant feature dimension and a variable number of task-irrelevant feature dimensions. Example cartoon ``daemon'' stimuli are depicted in Fig.~\ref{fig:psychology}c, where for instance horn height might be the distinguishing feature while colour, eye size, and mouth size might constitute task-irrelevant features. Without any irrelevant factors of variation ($\rho=1$), they report no curriculum benefit. By contrast when $75\%$ of features are irrelevant ($\rho=.25$), they record a strong curriculum effect, as shown in Fig.~\ref{fig:psychology}b bottom. This qualitative trend is also observed in our model (Fig.~\ref{fig:psychology}b top). While these experiments tested only two sparsity levels, further experiments could sample this dimension more extensively and test for interactions with the fraction of easy and hard examples. We note that while the connectionist literature has addressed the effect of curriculum in several settings \cite{plunkett1991rote,plunkett1991u,elman_learning_1993,krueger_flexible_2009}, we found that easy-to-hard effects appear even in a simple setup without need for complex networks and/or dynamics.

Finally, our results may shed light on self-generated curricula during human development \cite{raz_how_2019,smith_developmental_2017}. Children undergo a vocabulary spurt that coincides with their ability to grasp and centre objects in the visual field \cite{smith_developmental_2017}. Quantitative estimates of the amount of clutter (irrelevant objects) in self-generated views decrease due to this grasping ability, yielding a self-generated curriculum \cite{clerkin_real-world_2017,yu_embodied_2012}. Our model similarly predicts that reducing clutter should improve learning speed and performance. 

\begin{figure}
	\centering
        \includegraphics[width=0.58\linewidth]{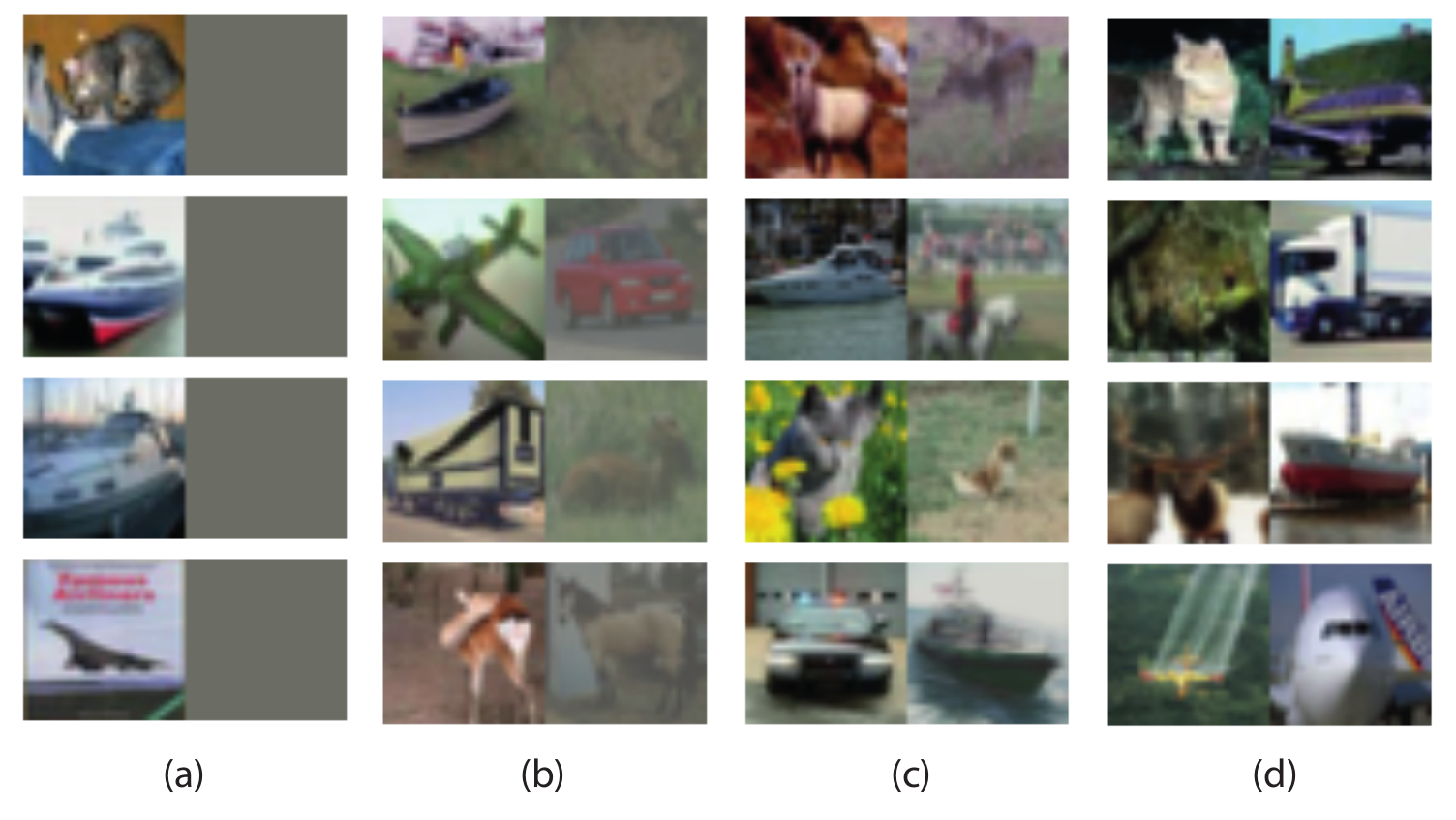}
        \includegraphics[width=0.41\linewidth]{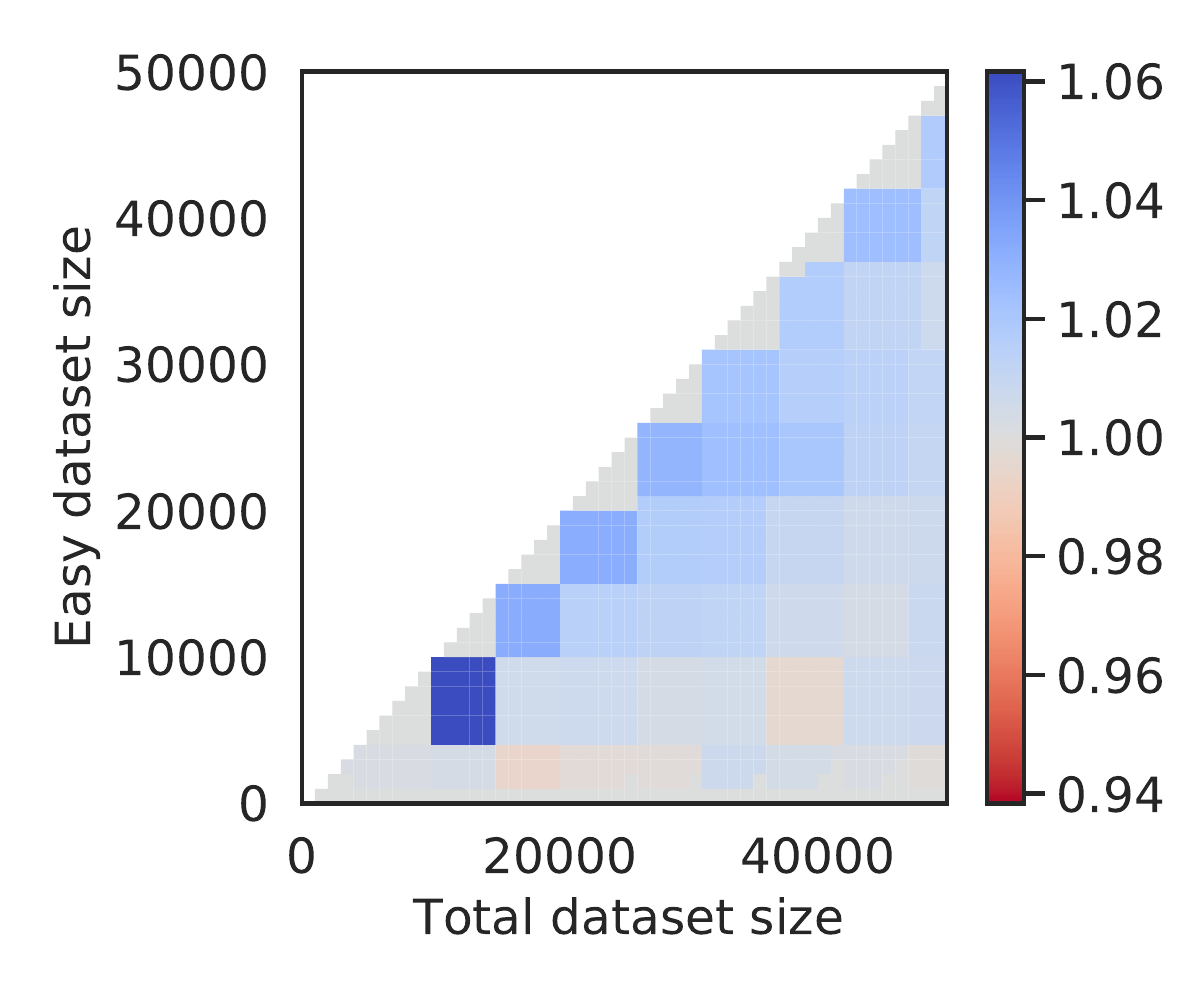}
	\caption{\textbf{Experimental setting on CIFAR10-derived data.}
	(a) Input samples combine a task-relevant image with a distractor image, and become progressively harder from left to right. 
	(b) Ratio between final accuracy on hard instances for curriculum learning versus no curriculum.
	$\eta, \gamma, \gamma_{12},$ init, and stopping time are optimised. 
	}
	\vspace{-1mm}
	\label{fig:real_data}
\end{figure}

\textbf{Real-World Demonstration.} To verify this prediction in a richer visual setting, we construct a simple cluttered object classification task from the CIFAR10 dataset \cite{krizhevsky_learning_nodate} by patching two images together into a $32\times64$ input image (Fig.~\ref{fig:real_data}a). The task is to produce the class label of the image on the left. The right image is a distractor that is irrelevant to the classification. 
To vary difficulty, we scale the contrast of the irrelevant image (Fig.~\ref{fig:real_data}a-d). 
We train a single-layer network with the cross-entropy loss and the curriculum protocol with Gaussian prior between two curriculum stages, implemented in Pytorch Lightning 
to ensure that training parameters accord with standard practice. 
We optimised hyperparameters in each curriculum phase separately.
We trained all combinations of five elastic penalties log spaced between $1e-3$ and $1e2$, and weight decay parameters $\{0, .2, .5\}$. We then compute the best performing model for five random seeds and take the mean over seeds. Further dataset, model and experimental details are given in Appendix \ref{sec:cifar}. As shown in Fig.~\ref{fig:real_data}b, curriculum improves performance, particularly when easy examples make up a large proportion of the dataset, confirming that curricula that reduce clutter can benefit learning.

\section{Conclusions}

We analysed a model of curriculum learning introduced by \cite{bengio2009curriculum} and amenable of analytical treatment. This simple setting sheds light on results observed in the cognitive science and machine learning literature, and the theoretical tractability allows for exploration of a wide range of parameters that would be costly to obtain through experiments. Future work will need to move beyond models with simple loss landscapes to address the impact of curricula in complex tasks like reinforcement learning. Nevertheless, the model recapitulates a variety of observations in the literature \cite{liu2008easy,orduna2012evoked,church2013temporal}, revealing that easy-to-hard effects can appear when a sparse signal is embedded in many irrelevant dimensions of variation. 
We find that making the algorithm curriculum-aware by modifying the loss can better exploit curricula, offering a potential route for improved practical algorithms. Other curriculum-aware approaches are possible such as adapting the learning algorithm \cite{ruiz2021tilting} or the architecture \cite{elman_learning_1993}.
On the psychology side, our predictions can help in designing new experiments, for instance testing the counter-intuitive benefit of anti-curriculum learning for intermediate sparsity.

\begin{ack}
    We thank Miguel Ruiz-Garcia and Ronald Dekker for important discussions.
    L.S. acknowledges funding from the ERC European Union Horizon 2020 Research and Innovation Program Grant Agreement 714608-SIiLe. S.S.M. \& A.S. were supported by a Wellcome and Royal Society Henry Dale Fellowship (216386/Z/19/Z) and Sainsbury Wellcome Centre Core Grant (219627/Z/19/Z, GAT3755). A.S. is a CIFAR Azrieli Global Scholar in the Learning in Machines \& Brains programme.
\end{ack}


%% file: camera_ready_ap.tex
\appendix
\numberwithin{equation}{section}
\numberwithin{figure}{section}

\section{State evolution of the online dynamics}

In this section we show how to derive the dynamical equations for the online dynamics. 
The equations given in an implicit form in the main text, $f_{Q_r}, f_{Q_i}, f_R$, are reported explicitly at the end of the next section, Eqs.~(\ref{eq:app:fQr}-\ref{eq:app:fR}).  
Finally, in the subsequent section, we comment on how the state evolution is modified to deal with the Gaussian priors and we derive the new dynamical equations for that case. 

\paragraph*{Derivation}

We follow the derivation proposed in \cite{biehl1995learning,saad1995exact} to derive the averaged high-dimensional dynamical equations. The student is a 1-layer network that minimises sample-wise the square error
\begin{equation}
	\mathcal L^\mu = \frac12\left(y^\mu-\hat y^\mu\right)^2 \doteq \frac12\left(\delta^\mu\right)^2.
\end{equation}
Given $\phi(\cdot)=\text{sign}(\cdot), \sigma(\cdot)=\text{erf}(\cdot/\sqrt2)$,
the online stochastic gradient descent updates are
\begin{align}
    \label{eq:app:dyn}
	& \pmb W^{\mu+1} = \pmb W^\mu - \frac{\eta}{\sqrt{N}} \sigma'(\lambda^\mu_r + \lambda^\mu_i)\delta^\mu \pmb x^\mu,
\end{align}
with
\begin{align}
	& \lambda^\mu_r = \frac1{\sqrt{N}} \pmb W_r \cdot \pmb x^\mu_r,
	\\
	& \lambda^\mu_i = \frac1{\sqrt{N}} \pmb W_i \cdot \pmb x^\mu_i,
	\\
	& \rho^\mu = \frac1{\sqrt{N}} \pmb W_T \cdot \pmb x^\mu_r.
\end{align}

The evolution of the dynamics can be tracked using 4 order parameters:
\begin{align}
	& Q_r = \frac1N \pmb W_r \cdot \pmb W_r,
	\\
	& Q_i = \frac1N \pmb W_i \cdot \pmb W_i,
	\\
	& R = \frac1N \pmb W_r \cdot \pmb W_T,
	\\
	& T = \frac1N \pmb W_T \cdot \pmb W_T;
\end{align}
representing the overlaps between the weights of student (relevant and irrelevant parts) and teacher.

The evolution of those follow from the definition of the dynamics Eq.~\eqref{eq:app:dyn}. In the high-dimensional limit the random variables in the problem concentrates around the mean, therefor to the leading order we have the following equations
\begin{align}
	\label{eq:oSGD_Qr}
	& Q_r[k+1] = Q_r[k] + \frac1N\left[ 2\eta \mathbb E[ \delta\ \sigma'(\lambda_r+\lambda_i) \lambda_r ] + \rho\Delta \eta^2 \mathbb E[ \delta^2\ \sigma'(\lambda_r+\lambda_i)^2 ] \right];
	\\
	\label{eq:oSGD_Qi}
	& Q_i[k+1] = Q_r[k] + \frac1N\left[ 2\eta \mathbb E[ \delta\ \sigma'(\lambda_r+\lambda_i) \lambda_i ] + (1-\rho)\Delta  \eta^2 \mathbb E[ \delta^2\ \sigma'(\lambda_r+\lambda_i)^2 ] \right];
	\\
	\label{eq:oSGD_R}
	& T[k+1] = Q_r[k] + \frac1N\left[ \eta \mathbb E[ \delta\ \sigma'(\lambda_r+\lambda_i) \rho ] \right].
	\\
\end{align}
Where the expectation acts with respect to all the stochastic variables.
In order to obtain explicit formulae we need to evaluate those averages. The random variables in the equations -- $\lambda_r$, $\lambda_i$ and $\rho$ -- are Gaussian with zero mean, to characterise them we only need their covariance:   
$$
\Sigma_{\lambda_r,\lambda_i,\rho} = \begin{pmatrix}
	Q_r & 0 & R \\
	0 & Q_i & 0 \\
	R & 0 & T
\end{pmatrix}.
$$

In order to derive analytical expression we must evaluate the expected values: $\mathbb E[\phi(\rho)\sigma'(\lambda)\rho]$, $\mathbb E[\phi(\rho)\sigma'(\lambda)\lambda]$, $\mathbb E[\sigma(\lambda)\sigma'(\lambda)\rho]$, $\mathbb E[\sigma(\lambda)\sigma'(\lambda)\lambda]$, $\mathbb E[\phi(\rho)^2\sigma'(\lambda)^2]$, $\mathbb E[\sigma(\lambda)^2\sigma'(\lambda)^2]$, and $\mathbb E[\phi(\rho)\sigma(\lambda)\sigma'(\lambda)^2]$. Where $\sigma$ is the activation function of the student and $\phi$ is the activation function of the teacher (in particular $\phi(\cdot)=\text{sign}(\cdot)$ for classification).


\begin{align}
	\begin{split}
		& \mathbb E[\phi(\rho)\sigma'(\lambda)\rho] 
		= \frac2\pi\frac{\sqrt{T(Q_r+Q_i+1)-R^2}}{Q_r+Q_i+1}
	\end{split}
	\\
	& \mathbb E[\phi(\rho)\sigma'(\lambda)\lambda_r] = \frac2\pi \frac{R(Q_i+1)}{Q_r+Q_i+1} \frac1{\sqrt{T(Q_r+Q_i+1)+R^2}}.
	\\
	& \mathbb E[\phi(\rho)\sigma'(\lambda)\lambda_i] = -\frac2\pi \frac{R Q_i}{Q_r+Q_i+1} \frac1{\sqrt{T(Q_r+Q_i+1)+R^2}}.
	\\
	& \mathbb E[\sigma(\lambda)\sigma'(\lambda)\rho] = \frac2\pi \frac{R}{Q_r+Q_i+1} \sqrt{\frac{Q_i+1}{2Q_i^2+2Q_rQ_i+3Q_i+2Q_r+1}}.
	\\
	& \mathbb E[\sigma(\lambda)\sigma'(\lambda)\lambda_r] = \frac2\pi \frac{Q_r}{Q_r+Q_i+1} \sqrt{\frac{Q_i+1}{2Q_i^2+2Q_rQ_i+3Q_i+2Q_r+1}}.
	\\
	& \mathbb E[\sigma(\lambda)\sigma'(\lambda)\lambda_i] = \frac2\pi \frac{Q_i}{Q_r+Q_i+1} \sqrt{\frac{Q_r+1}{2Q_r^2+2Q_rQ_i+3Q_r+2Q_i+1}}.
	\\
	\begin{split}
		& \mathbb E[\phi(\rho)^2\sigma'(\lambda)^2] = \frac2\pi\frac1{\sqrt{2Q_r+2Q_i+1}}.
	\end{split}
	\\
	& \mathbb E[\sigma(\lambda)^2\sigma'(\lambda)^2] = \frac4{\pi^2}\frac1{\sqrt{1+2(Q_r+Q_i)}}\sin^{-1}\left(\frac{Q_r+Q_i}{1+3(Q_r+Q_i)}\right).
	\\
	\begin{split}
		& \mathbb E[\phi(\rho)\sigma(\lambda)\sigma'(\lambda)^2] = \frac4{\pi^2}\frac1{\sqrt{2(Q_r+Q_i)+1}}
		\\
		&\quad\quad
		\sin^{-1}\left(
	\frac{R\sqrt{Q_r+Q_i}}{\sqrt{3(Q_r+Q_i)+1}\sqrt{(2Q_r+2Q_i+1)[T(Q_r+Q_i)-R^2]+R^2}}\right).
	\end{split}
\end{align}

Finally, we can substitute those equations into the Eqs.~(\ref{eq:oSGD_Qr}-\ref{eq:oSGD_R}) and obtained the state evolution equations used in the main Sec.~\ref{sec:online_dyn}:
\begin{align}
    \label{eq:app:fQr}
    \begin{split}
        & f_{Q_r}\big(Q_r[k],Q_i[k],R[k],T\big) = (1-\eta\gamma)^2Q_r[k] + 
        \frac{4\eta (1-\eta\gamma)}{N\pi (Q_r[k]+\Delta Q_i[k]+1)} \times 
        \\
        &\quad\quad 
        \Bigg[ \frac{R[k](\Delta Q_i[k]+1)}{\sqrt{T(Q_r[k]+\Delta Q_i[k]+1)+R[k]^2}} -
        \frac{Q_r[k]}{\sqrt{2Q_r[k]+2\Delta Q_i[k]+1}}\Bigg]
        \\
        & \quad\quad + \frac4{\pi^2}\frac{\rho\eta^2}{N\sqrt{2(Q_r[k]+\Delta Q_i[k])+1}} \Bigg[ \frac{\pi}2+ \sin^{-1}\left(\frac{Q_r[k]+\Delta Q_i[k]}{1+3(Q_r[k]+\Delta Q_i[k])}\right) + 
		\\
		&\quad\quad
		- 2\sin^{-1}\left(
	\frac{R[k]}{\sqrt{3(Q_r[k]+\Delta Q_i[k])+1}\sqrt{T(2Q_r[k]+2\Delta Q_i[k]+1)-2R[k]^2}}\right) \Bigg];
    \end{split}
    \\
    \begin{split}
        & f_{Q_i}\big(Q_r[k],Q_i[k],R[k],T\big) = (1-\eta\gamma)^2Q_i[k] - 
        \frac{4\eta (1-\eta\gamma) \Delta Q_i[k]}{N\pi (Q_r[k]+\Delta Q_i[k]+1)} \times
        \\
        &\quad\quad
        \Bigg[
        \frac{R[k]}{\sqrt{T(Q_r[k]+\Delta Q_i[k]+1)+R[k]^2}} + 
        \frac1{\sqrt{2Q_r[k]+2\Delta Q_i[k]+1}}\Bigg] +
        \\
        & \quad\quad + \frac4{\pi^2}\frac{(1-\rho)\Delta\eta^2}{N\sqrt{2(Q_r[k]+\Delta Q_i[k])+1}} \Bigg[ \frac{\pi}2+ \sin^{-1}\left(\frac{Q_r[k]+\Delta Q_i[k]}{1+3(Q_r[k]+\Delta Q_i[k])}\right) + 
		\\
		&\quad\quad
		- 2\sin^{-1}\left(
	\frac{R[k]}{\sqrt{3(Q_r[k]+\Delta Q_i[k])+1}\sqrt{T(2Q_r[k]+2\Delta Q_i[k]+1)-2R[k]^2}}\right) \Bigg];
    \end{split}
    \\
    \label{eq:app:fR}
    \begin{split}
    & f_{R}\big(Q_r[k],Q_i[k],R[k],T\big) = (1-\eta\gamma)R[k] + \frac{2\eta}{N\pi(Q_r[k]+\Delta Q_i[k]+1)}
        \times
        \\
        &\quad\quad 
		\Bigg[  \frac{T(Q_r[k]+\Delta Q_i[k]+1)-R[k]^2}{\sqrt{T(Q_r[k]+\Delta Q_i[k]+1)-R[k]^2}}
		- \frac{R[k]}{\sqrt{2Q_r[k]+2\Delta Q_i[k]+1}}\Bigg].
    \end{split}
\end{align}

\paragraph*{Elastic coupling}

\begin{figure}
    \centering
	\begin{subfigure}[b]{\linewidth}
        \centering
        \includegraphics[width=\linewidth]{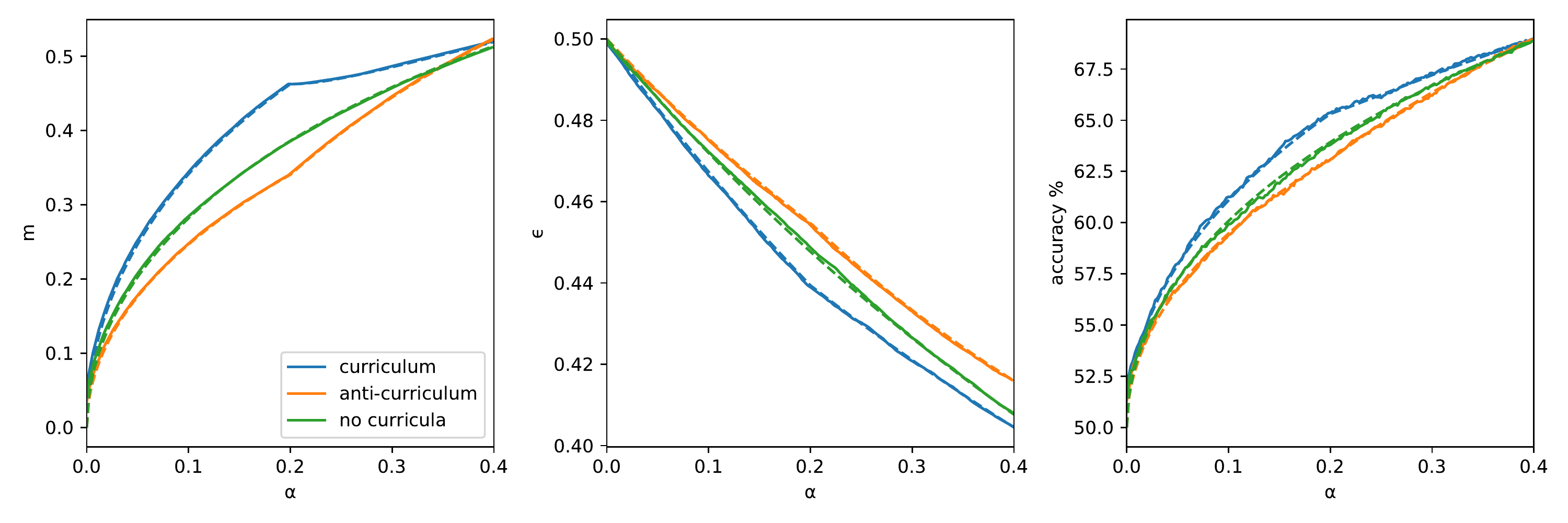}
		\caption{No elastic coupling.}
    \end{subfigure}
    \centering
	\begin{subfigure}[b]{\linewidth}
        \centering
        \includegraphics[width=\linewidth]{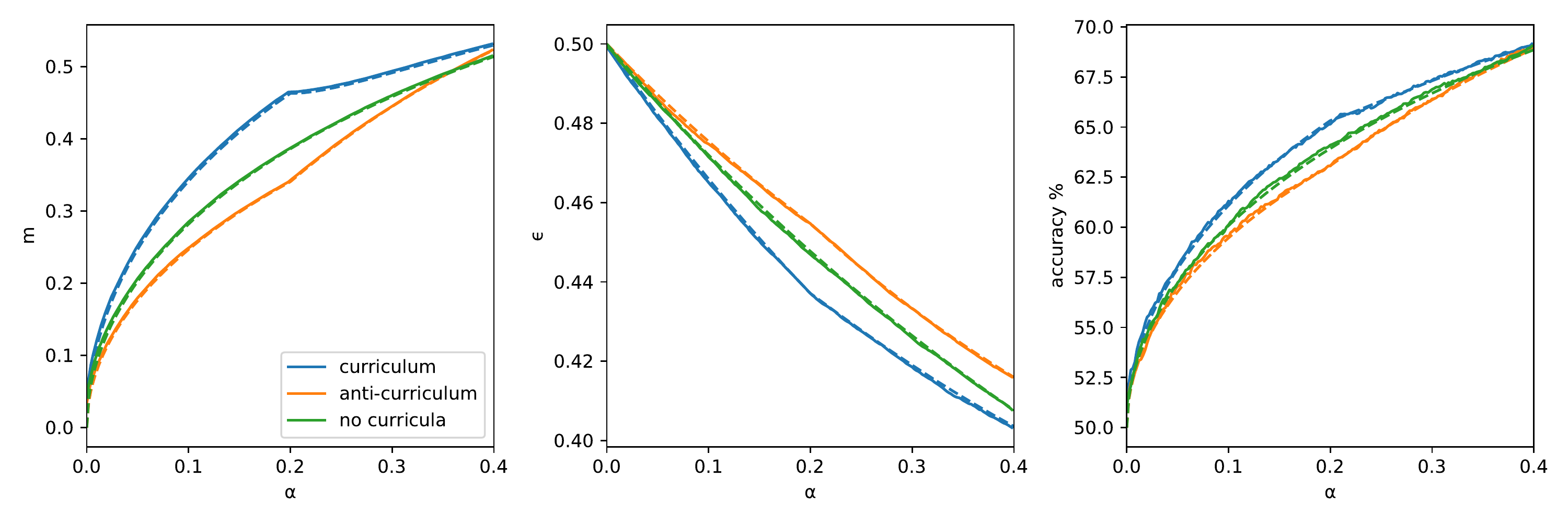}
		\caption{Optimal elastic coupling.}
    \end{subfigure}
    \caption{\textbf{Effect of elastic coupling in the curriculum.} Figures showing the teacher-student cosine, the validation loss, and the accuracy of the three learning strategies. The two figures show the performance in presence (above) and absence (below) of elastic coupling. The dashed lines are obtained from the theoretical analysis, the full line come from the average of 500 simulations. The parameters $\eta$, $\gamma$, initialisation are set to the optimal values for each protocol. Parameters: $\rho=0.5$, $\alpha_1=0.2$, $\alpha_2=0.2$, $\Delta_1=0$, $\Delta_2=1$.
    \label{fig:sm:elastic}
    }
\end{figure}

The introduction of the elastic coupling between stages of learning adds five new order parameters: three of them are just reminder of the previous stage and do not need to by updated  $\tilde Q_r = \pmb W^r_1\cdot \pmb W^r_1/N$, $\tilde Q_i = \pmb W^i_1\cdot \pmb W^i_1/N$, and $\tilde R = \pmb W^i_1\cdot \pmb W^T/N$; two measure the correlation between the two stages $S_r = \pmb W^r_1\cdot \pmb W^r_2/N$ and $S_i = \pmb W^i_1\cdot \pmb W^i_2/N$ to the equations. These terms have associated their own state evolution equations slightly modified the updates of the other order parameters.

\begin{align}
	\label{eq:oSGD_Qr_elastic}
	\begin{split}
		& Q_r[k+1] = (1-\eta\gamma+\eta\gamma_{12})^2 Q_r[k] + \frac{2\eta}N (1-\eta\gamma+\eta\gamma_{12}) \mathbb{E}[ \delta\ \sigma'(\lambda_r+\lambda_i) \lambda_r ]
		\\
		& \quad\quad + \rho\Delta \frac{\eta^2}N \mathbb{E}[ \delta^2\ \sigma'(\lambda_r+\lambda_i)^2 ] + 2\eta \gamma_{12} (1-\eta\gamma+\eta\gamma_{12}) S_r[k] + \eta^2 \gamma_{12}^2 \tilde Q_r[k]
		\\
		& \quad\quad - \frac{2 \eta^2 \gamma_{12}}N \mathbb{E}[ \delta\ \sigma'(\lambda_r+\lambda_i) \tilde\lambda_r ];
	\end{split}
	\\
	\label{eq:oSGD_Qi_elastic}
	\begin{split}
		& Q_i[k+1] = (1-\eta\gamma+\eta\gamma_{12})^2 Q_i[k] + \frac{2\eta}N (1-\eta\gamma+\eta\gamma_{12}) \mathbb{E}[ \delta\ \sigma'(\lambda_r+\lambda_i) \lambda_i ]
		\\
		& \quad\quad + (1-\rho)\Delta  \frac{\eta^2}N \mathbb{E}[ \delta^2\ \sigma'(\lambda_r+\lambda_i)^2 ] + 2\eta \gamma_{12} (1-\eta\gamma+\eta\gamma_{12}) S_i[k]
		\\
		& \quad\quad + \eta^2 \gamma_{12}^2 \tilde Q_i[k] - \frac{2 \eta^2 \gamma_{12}}N \mathbb{E}[ \delta\ \sigma'(\lambda_r+\lambda_i) \tilde\lambda_i ];
	\end{split}
	\\
	\label{eq:oSGD_R_elastic}
	& R[k+1] = (1-\eta\gamma+\eta\gamma_{12}) R[k] + \frac{\eta}N \mathbb{E}[ \delta\ \sigma'(\lambda_r+\lambda_i) \rho ] -\eta\gamma_{12} \tilde R[k];
	\\
	\label{eq:oSGD_R12r_elastic}
	& S_r[k+1] = (1-\eta\gamma+\eta\gamma_{12}) S_r[k] + \frac{\eta}N \mathbb{E}[ \delta\ \sigma'(\lambda_r+\lambda_i) \tilde\lambda_r ] -\eta\gamma_{12} \tilde Q_r[k];
	\\
	\label{eq:oSGD_R12i_elastic}
	& S_i[k+1] = (1-\eta\gamma+\eta\gamma_{12}) S_i[k] + \frac{\eta}N \mathbb{E}[ \delta\ \sigma'(\lambda_r+\lambda_i) \tilde\lambda_i ] -\eta\gamma_{12} \tilde Q_i[k].
\end{align}
Introduced $\tilde\lambda_r = \frac1{\sqrt{N}}\pmb x_r\cdot \tilde{\pmb W}_r$ and $\tilde\lambda_i = \frac1{\sqrt{N}}\pmb x_i\cdot \tilde{\pmb W}_i$, this two additional random variables need to be averaged together with the others. The joint distribution of $\lambda_r, \lambda_i, \tilde\lambda_r, \tilde\lambda_i,\rho$ is still Gaussian with zero mean and covariance
$$
\Sigma_{\lambda_r, \lambda_i, \tilde\lambda_r, \tilde\lambda_i,\rho} = \begin{pmatrix}
	Q_r & 0 & \tilde S_r & 0 & R \\
	0 & Q_i & 0 & \tilde S_i & 0 \\
	\tilde S_r & 0 & \tilde Q_r & 0 & \tilde R\\
	0 & \tilde S_i & 0 & \tilde Q_i & 0 \\
	R & 0 & \tilde R & 0 & T
\end{pmatrix}.
$$
Notice that, a part from a slight change of the existing equations, the coupling introduces only two additional integrals $\mathbb{E}[ \delta\ \sigma'(\lambda_r+\lambda_i) \tilde\lambda_r ]$ and $\mathbb{E}[ \delta\ \sigma'(\lambda_r+\lambda_i) \tilde\lambda_i ]$.
After long, but straightforward, computations we obtain
\begin{align}
	\begin{split}
		& \mathbb{E}[ \delta\ \sigma'(\lambda_r+\lambda_i) \tilde\lambda_r ] = \frac2\pi \frac{S_r}{Q_r+Q_i+1}\frac{Q_i+1}{2Q_i^2+2Q_rQ_i+3Q_i+2Q_r+1} + 
		\\
		& \quad\quad - \frac2\pi \frac{TS_r-R\tilde{R}}{Q_r T -R^2} \frac{R(Q_i+1)}{Q_r+Q_i+1}\frac1{\sqrt{T(Q_r+Q_i+1)-R^2}} +
		\\
		& \quad\quad - \frac2\pi \frac{T\tilde{R}-R S_r}{Q_r T -R^2} \frac1{\sqrt{T(Q_r+Q_i+1)-R^2}}\frac1{\frac1T+\frac{R^2}{Q_rT-R^2}\left(\frac1T-\frac{Q_i+1}{T(Q_r+Q_i+1)-R^2}\right)},
	\end{split}
	\\
	\begin{split}
		& \mathbb{E}[ \delta\ \sigma'(\lambda_r+\lambda_i) \tilde\lambda_i ] = \frac2\pi \frac{S_i}{Q_r+Q_i+1}\frac{Q_r+1}{2Q_r^2+2Q_rQ_i+3Q_r+2Q_i+1} + 
		\\
		& \quad\quad - \frac2\pi \frac{S_i R}{Q_r+Q_i+1} \frac1{\sqrt{T(Q_r+Q_i+1)-R^2}}.
	\end{split}
\end{align}
Finally all the expected values are known and we can obtain the analytic updates Eqs.~(\ref{eq:oSGD_Qr_elastic}-\ref{eq:oSGD_R12i_elastic}) with the coupling. Fig.~\ref{fig:sm:elastic}a shows an instance of the problem at $\alpha_1=0.2$ and $\alpha_2=0.2$, a situation that is particularly adversarial for curriculum according the phase diagram Fig.~\ref{fig:classification_curricula_comparison_heatmap}. This situation is treated by the introduction of Gaussian priors, Fig.~\ref{fig:sm:elastic}b, consistently with the phase diagram in Fig.~\ref{fig:real_data}c.

\section{Replica computation for the batch case}

We here the detailed replica computation employed to obtain the analytic description of curriculum learning in the batch case, in section 4. As mentioned in the main, we aim to study a coupled system, represented by the following partition function:
\begin{equation}
    \left< Z(\pmb{W}_2, \pmb{W}_1; \mathcal{D}_1, \mathcal{D}_2) \right>_{\pmb{W}_1} = \int d\pmb{W}_1 \frac{e^{-\beta_1 \mathcal{L}_{\gamma_1}(\pmb{W}_1, \mathcal{D}_1)}}{Z_1(\pmb{W}_1)} \log \int d\pmb{W}_2 \,e^{-\beta_2 \left( \mathcal{L}_{\gamma_2}(\pmb{W}_2,\mathcal{D}_2) + \frac{\gamma_{12}}{2} \lVert \pmb{W}_2 - \pmb{W}_1 \rVert^2_2 \right)},
\end{equation}
where the examples is $\mathcal{D}_1,\mathcal{D}_2$ are characterised by a different variances in the irrelevant components. 

This type of quantity is usually denoted as a ``disordered'' partition function in statistical physics jargon, meaning that it is still dependent on a given realisation of the datasets -- i.e., the source of disorder in this model. We want to characterise a typical realisation of this object, in the high-dimensional limit. However, because of its long-tailed statistics, the partition function turns out not to be a self-averaging quantity, i.e. its expectation over the dataset realisations will not correspond to the typical case scenario we are after. It is instead better to focus on the computation of the associated average free-entropy:
\begin{equation}
    \Phi = \lim_{N\to\infty}\lim_{\beta_1,\beta_2\to\infty} \frac{1}{\beta_2 N} \left< \log \left< Z(\pmb{W}_2, \pmb{W}_1; \mathcal{D}_1, \mathcal{D}_2) \right>_{\pmb{W}_1} \right>_{\mathcal{D}_1, \mathcal{D}_2}.
\end{equation}
What is immediately apparent is that we have to take the expectation of a logarithm, which is not tractable with rigorous mathematical methods. Moreover, we also have to average over the measure for $\pmb W_1$, which is also a complicated operation. 

Fortunately, replica theory offers a method for approaching this calculation \cite{franz1997phase, saglietti2020solvable}. The idea is to exploit two separate replica tricks: 
\begin{itemize}
    \item in order to evaluate the disorder average, the logarithm can be removed by replicating the second weight configuration, i.e. introducing $n$ identical replicas $\{ \boldsymbol{W}^a_2\}_{a=1}^n$, and extrapolating the final result from the $n\to 0$ limit. This is based on the mathematical identity $\log x = \lim_{n\to 0} \partial_n x^n $. 
    \item the average over the teacher can instead be computed by introducing $\tilde{n}-1$ non-interacting and a single interacting replica of the first weight configuration $\{ {\boldsymbol{w}}^c_1\}_{c=1}^{\tilde{n}}$. Thus, only the $c=1$ replica will enter the Gaussian prior in the student measure. The sought statistical average is again recovered in the limit $\tilde{n}\to 0$.
\end{itemize} 
Because of the high-dimensional limit we are considering, all typical realisations of the teacher vector with a given sparsity $\rho$ will yield an identical free-entropy. Thus, we can avoid averaging and instead fix a gauge $\pmb W_{T,i}=1$ for $i=1,\dots,\rho N$ and $\pmb W_{T,i}=0$ elsewhere.
In order to simplify the presentation, in the following we will assume that the datasets contain respectively $\alpha_1$ and $\alpha_2$ patterns, and that a curriculum ordering was employed, $\Delta_1<\Delta_2$.  Moreover, to avoid confusion with component and replica indices, we will denote with $\tilde{\pmb W} =\pmb W_1$ and ${\pmb W} =\pmb W_2$, so that all quantities with a tilde refer to the optimisation on the first dataset. 

After the described replication procedures, we get the following expression for the average free-entropy:
\small
\begin{equation}
\Phi=\frac{1}{N}\lim_{n,\tilde{n}\to0}\partial_{n}\left<\lim_{\tilde{\beta},\beta\to\infty}\frac{1}{\beta} \int\prod_{c=1}^{\tilde{n}}d\tilde{\boldsymbol{W}}^{c}e^{-\frac{\tilde{\beta}\gamma_1}{2}\lVert\tilde{\boldsymbol{W}}^{c}\rVert_{2}^{2}}\prod_{\mu=1}^{\alpha_{1} N}\prod_{c=1}^{\tilde{n}}e^{-\frac{\beta}{2}\ell\left(\mathrm{sign}\left(\sum_{i=1}^{\rho N}\frac{x_{i}^{\mu}}{\sqrt{N}}\right),\sigma\left(\sum_{i=1}^{N}\frac{\tilde{W}_{i}^{c}x_{i}^{\mu}\left(\Delta_{1}\right)}{\sqrt{N}}\right)\right)}\right.
\end{equation}
\[
\times\int\prod_{a=1}^{n}d\boldsymbol{W}^{a}e^{-\frac{\beta\gamma_2}{2}\lVert\boldsymbol{W}^{a}\rVert_{2}^{2}}e^{-\frac{\beta\gamma_{12}}{2}\lVert\boldsymbol{W}^{a}-\tilde{\boldsymbol{W}}^{1}\rVert_{2}^{2}} \left.\prod_{\mu=1}^{\alpha_{2}}\prod_{a}e^{-\frac{\beta}{2}\ell\left(\mathrm{sign}\left(\sum_{i=1}^{\rho N}\frac{x_{i}^{\mu}}{\sqrt{N}}\right),\sigma\left(\sum_{i=1}^{N}\frac{W_{i}^{a}x_{i}^{\mu}\left(\Delta_{2}\right)}{\sqrt{N}}\right)\right)}\right>_{\left\{ \boldsymbol{x}^{\mu}\right\} },
\]
\normalsize
where $\ell(y, \hat{y})=\log(1+e^{-y \hat{y}})$ indicates the standard logistic loss. The next step is to explicitly compute the averages over the dataset realisations. Before doing that, we need to isolate the dependence of our expression on the patterns, and we achieve this by introducing Dirac's $\delta$-functions for the pre-activations. We will use the integral representation of the $\delta$, with integration variables $u$ for the teacher preactivations $\lambda$ for the student preactivations:
\small
\begin{equation}
\frac{1}{N}\lim_{n,\tilde{n}\to0}\partial_{n} \int\prod_{c=1}^{\tilde{n}}d\tilde{\boldsymbol{W}}^{c}e^{-\frac{\beta\lambda}{2}\lVert\tilde{\boldsymbol{W}}^{c}\rVert_{2}^{2}}\int \prod_{a=1}^{\tilde{n}}d\boldsymbol{W}^{a} e^{-\frac{\beta\lambda}{2}\lVert\boldsymbol{W}^{a}\rVert_{2}^{2}}e^{-\frac{\beta\lambda_{12}}{2}\lVert\boldsymbol{W}^{a}-\tilde{\boldsymbol{W}}^{1}\rVert_{2}^{2}}
\end{equation}
\[
\times \left<\int\prod_{\mu}\frac{d\tilde{u}_{1\mu}d\hat{\tilde{u}}_{1\mu}}{2\pi}e^{i\hat{\tilde{u}}_{1\mu}\left(\tilde{u}_{1\mu}-\sum_{i=1}^{\rho N}\frac{\left(\tilde{x}_{1}\right)_{i}^{\mu}}{\sqrt{N}}\right)}\int\prod_{\mu,c}\frac{d\tilde{\lambda}_{1\mu}^{c}d\hat{\tilde{\lambda}}_{1\mu}^{c}}{2\pi}e^{i\hat{\tilde{\lambda}}_{1\mu}^{c}\left(\lambda_{1\mu}^{c}-\sum_{i=1}^{N}\frac{\tilde{W}_{i}^{c}\left(\tilde{x}_{1}\right)_{i}^{\mu}}{\sqrt{N}}\right)} \right.
\]
\[
\left.\times \int\prod_{\mu}\frac{du_{2\mu}d\hat{u}_{2\mu}}{2\pi}e^{i\hat{u}_{2\mu}\left(u_{2\mu}-\sum_{i=1}^{\rho N}\frac{\left(x_{2}\right)_{i}^{\mu}}{\sqrt{N}}\right)}\int\prod_{\mu,a}\frac{d\lambda_{2\mu}^{a}d\hat{\lambda}_{2\mu}^{a}}{2\pi}e^{i\hat{\lambda}_{2\mu}^{a}\left(\lambda_{2\mu}^{a}-\sum_{i=1}^{N}\frac{W_{i}^{a}\left(x_{2}\right)_{i}^{\mu}}{\sqrt{N}}\right)}\right>_{\left\{ \boldsymbol{x}^{\mu} \right\}}\
\]
\[
\times \prod_{\mu,c}e^{-\frac{\beta}{2}\ell\left(\mathrm{sign}\left(\tilde{u}_{1\mu}\right),\sigma\left(\tilde{\lambda}_{1\mu}^{c}\right)\right)}\prod_{\mu,a}e^{-\frac{\beta}{2}\ell\left(\mathrm{sign}\left(u_{2\mu}\right),\sigma\left(\lambda_{2\mu}^{a}\right)\right)}.
\]
\normalsize

Thus, the disorder average is now factorised and only involves exponential terms. Since the two datasets are independent now that we made the teacher explicit, we can take the averages over each one separately. In both cases we get:
\small
\begin{eqnarray}
 \left< .\right> & = & \prod_{i=1}^{\rho N}\mathbb{E}_{\left(x_{rel}\right)_{i}^{\mu}}e^{-i\left(\frac{\hat{u}}{\sqrt{N}}+\sum_{a}\hat{\lambda}_{a}^{\mu}\frac{W_{i}^{a}}{\sqrt{N}}\right)\left(x_{rel}\right)_{i}^{\mu}}\prod_{i=\rho N+1}^{N}\mathbb{E}_{\left(x_{irr}\right)_{i}^{\mu}}e^{-i\left(\sum_{a}\hat{\lambda}_{a}^{\mu}\frac{W_{i}^{a}}{\sqrt{N}}\right)\left(x_{irr}\right)_{i}^{\mu}}\nonumber\\
 & = & \prod_{i=1}^{\rho N}\left(1-i\left(\frac{\hat{u}}{\sqrt{N}}+\sum_{a}\hat{\lambda}_{a}^{\mu}\frac{W_{i}^{a}}{\sqrt{N}}\right)\overline{x_{rel}}-\frac{1}{2}\left(\frac{\hat{u}}{\sqrt{N}}+\sum_{a}\hat{\lambda}_{a}^{\mu}\frac{W_{i}^{a}}{\sqrt{N}}\right)^{2}Var\left(x_{rel}\right)\right)\nonumber\\ 
 && \times\prod_{i=\rho N+1}^{N}\left(1-i\sum_{a}\hat{\lambda}_{a}^{\mu}\frac{W_{i}^{a}}{\sqrt{N}}\overline{x_{irr}}-\frac{1}{2}\left(\sum_{a}\hat{\lambda}_{a}^{\mu}\frac{W_{i}^{a}}{\sqrt{N}}\right)^{2}Var(x_{irr})\right)\\
 & = & \prod_{i=1}^{\rho N}\left(1-\frac{1}{2N}\left(\hat{u}^{\mu}\right)^{2}-\frac{1}{N}\sum_{a}\hat{u}^{\mu}\hat{\lambda}_{a}^{\mu}W_{i}^{a}-\frac{1}{2N}\sum_{ab}\hat{\lambda}_{a}^{\mu}\hat{\lambda}_{b}^{\mu}W_{i}^{a}W_{i}^{b}\right)\prod_{i=\rho N+1}^{N}\left(1-\frac{\Delta^{\mu}}{2N}\sum_{ab}\hat{\lambda}_{a}^{\mu}\hat{\lambda}_{b}^{\mu}W_{i}^{a}W_{i}^{b}\right)\nonumber
\end{eqnarray}
\normalsize
\begin{equation}
    =  e^{-\frac{1}{2}\sum_{ab}\hat{\lambda}_{a}^{\mu}\hat{\lambda}_{b}^{\mu}\left(\frac{\sum_{i=1}^{\rho N}W_{i}^{a}W_{i}^{b}}{N}+\Delta\frac{\sum_{i=\rho N+1}^{N}W_{i}^{a}W_{i}^{b}}{N}\right)-\frac{\rho}{2}\left(\hat{u}^{\mu}\right)^{2}-\hat{u}^{\mu}\sum_{a}\hat{\lambda}_{a}^{\mu}\frac{\sum_{i=1}^{\eta N}W_{i}^{a}}{N}}.
\end{equation}
This expression suggests what are the order parameters that capture the interactions of the model, namely:
\begin{itemize}
\item the teacher-student overlap at the end of the first learning phase: $\tilde{R}^{c}=\frac{\sum_{i=1}^{\rho N}\tilde{W}_{i}^{c}}{N}$.
\item the teacher-student overlap at the end of the second learning phase: $R^{a}=\frac{\sum_{i=1}^{\rho N}W_{i}^{a}}{N}$
\item the norm of the student after the first stage, decomposed into relevant/irrelevant parts: $\tilde{Q}_{r}^{cd}=\frac{\sum_{i=1}^{\rho N}\tilde{W}_{i}^{c}\tilde{W}_{i}^{d}}{N}$,
$\tilde{Q}_{i}^{cd}=\frac{\sum_{i=\rho N+1}^{N}\tilde{W}_{i}^{c}\tilde{W}_{i}^{d}}{N}$
\item the norm of the student after the second stage, decomposed into relevant/irrelevant parts: $Q_{r}^{ab}=\frac{\sum_{i=1}^{\rho N}W_{i}^{a}W_{i}^{b}}{N}$, $Q_{i}^{ab}=\frac{\sum_{i=\rho N+1}^{N}W_{i}^{a}W_{i}^{b}}{N}$
\end{itemize}
Therefore, after introducing these definitions by means of Dirac's $\delta$-functions, we can rewrite our replicated expression as:
\small
\[
\Omega^{n}=\int\prod_{c}\frac{d\tilde{R}^{c}d\hat{\tilde{R}}^{c}}{2\pi/N}\int\prod_{a}\frac{dR^{a}d\hat{R}^{a}}{2\pi/N}\int\prod_{cd}\frac{d\tilde{Q}_{r}^{cd}d\hat{\tilde{Q}}_{r}^{cd}}{2\pi/N}\int\prod_{cd}\frac{d\tilde{Q}_{i}^{cd}d\hat{\tilde{Q}}_{i}^{cd}}{2\pi/N}\int\prod_{ab}\frac{dQ_{r}^{ab}d\hat{Q}_{r}^{ab}}{2\pi/N}\int\prod_{ab}\frac{dQ_{i}^{ab}d\hat{Q}_{i}^{ab}}{2\pi/N}
\]
\begin{equation}
\times G_{i}\,G_{S}\left(\hat{\tilde{R}},\hat{R},\hat{\tilde{Q}}_{r},\hat{Q}_{r}\right)^{\rho N}G_{S}\left(0,0,\tilde{Q}_{i},Q_{i}\right)^{\left(1-\rho\right)N} G_{E}\left(\Delta_1,\tilde{Q}_{r},\tilde{Q}_{i},\tilde{R},\tilde{n}\right)^{\alpha_{1}N} G_{E}\left(\Delta_2,Q_{r},Q_{i},R,n\right)^{\alpha_{2} N}
\end{equation}
\normalsize
Where we introduced interaction, entropic and energetic potentials:
\begin{equation}
G_{i}=\exp\left(-N\left(\sum_{c}\hat{\tilde{m}}^{c}\tilde{m}^{c}+\sum_{a}\hat{m}^{a}m^{a}+\sum_{cd}\hat{\tilde{Q}}_{r}^{cd}\tilde{Q}_{r}^{cd}+\sum_{cd}\hat{\tilde{Q}}_{i}^{cd}\tilde{Q}_{i}^{cd}+\sum_{ab}\hat{Q}_{r}^{ab}Q_{r}^{ab}+\sum_{ab}\hat{Q}_{i}^{ab}Q_{i}^{ab}\right)\right)
\end{equation}
\begin{equation}
G_{S}\left(\tilde{R},R,\tilde{Q},Q\right)=\int\prod_{c}\left[d\tilde{W}^{c}e^{-\frac{\beta\gamma}{2}\left(\tilde{W}^{c}\right){}^{2}}\right]e^{-\frac{n\beta\gamma_{12}}{2}(\tilde{W}^{1})^{2}}\int\prod_{a}\left[dW^{a}e^{-\frac{\beta\left(\gamma+\gamma_{12}\right)}{2}\left(W^{a}\right)^{2}}\right]
\end{equation}
\[
\times \exp\left(\sum_{c}\hat{\tilde{R}}^{c}\tilde{W}^{c}+\sum_{a}\hat{R}^{a}W^{a}+\sum_{cd}\hat{\tilde{Q}}^{cd}\tilde{W}^{c}\tilde{W}^{d}+\sum_{ab}\hat{Q}^{ab}W^{a}W^{b}+\beta\gamma_{12}W^{a}\tilde{W}^{1}\right)
\]

\begin{equation}
G_{E}\left(\Delta,Q_{r},Q_{i},m,n\right)=\int\frac{dud\hat{u}}{2\pi}e^{iu\hat{u}}e^{-\frac{\rho}{2}\left(\hat{u}\right)^{2}}\int\prod_{a=1}^{n}\frac{d\lambda^{a}d\hat{\lambda}^{a}}{2\pi}e^{i\lambda^{a}\hat{\lambda}^{a}}
\end{equation}
\[
\times e^{-\frac{1}{2}\sum_{ab}\hat{\lambda}_{a}\hat{\lambda}_{b}\left(Q_{r}^{ab}+\Delta Q_{i}^{ab}\right)-\hat{u}\sum_{a}\hat{\lambda}_{a}R^{a}-\frac{\beta}{2}\ell\left(u,\lambda^{a}\right)}
\]

\subsubsection*{Replica Symmetric Ansatz}
The replica trick allowed us to express the average free-entropy as a function of the overlap order parameters. However, these objects are $n\times n$ matrices or $n$-dimensional vectors and in principle we have to average over all their possible realisations. Fortunately, the integrand function is exponential in $N$ and in the thermodynamic limit $N\to\infty$ the integrals are dominated by the extremisers of the action, and thus can be approximated with the saddle-point method. Still, we need a guess for how to parametrise these order parameters. The simplest possible ansatz, which turns out to be the correct one in convex problems as the one at hand, is the so-called Replica Symmetric ansatz, given by:
\begin{itemize}
\item $\tilde{R}^{c}=\tilde{R}$
\item $R^{a}=R$
\item $\tilde{Q}_{r/i}^{cd}=\tilde{q}_{r/i}$, for $c\neq d$; $\tilde{Q}_{r/i}^{cd}=\tilde{Q}_{r/i}$
for $c=d$.
\item $Q_{r/n}^{ab}=q_{r/n}$ for $a\neq b$; $Q_{r/n}^{ab}=Q_{r/n}$
for $a=b$.
\end{itemize}
We also perform a Wick rotation $-i\hat{Q}_{ac,bd}\to\hat{Q}_{ac,bd}$ in order to deal with real valued conjugate parameters and pose a similar ansatz for them. In the next paragraph we will compute the three terms separately, and finally put them together in the expression for the RS free-entropy.

\subsubsection*{Interaction term}
We start by evaluating the interaction term, or better its normalised logarithm $g_i = \lim_{\tilde{n}\to0} \log G_i / (nN)$:
\begin{eqnarray}
g_{i} &=& -\lim_{\tilde{n}\to0}\frac{1}{n}\left(\tilde{n}\hat{\tilde{R}}\tilde{R}+n\hat{R}R+\tilde{n}\left(\frac{\hat{\tilde{Q}}_{r}\tilde{Q}_{r}}{2}+\frac{\hat{\tilde{Q}}_{i}\tilde{Q}_{i}}{2}\right)+\frac{\tilde{n}(\tilde{n}-1)}{2}\left(\hat{\tilde{q}}_{r}\tilde{q}_{r}+\hat{\tilde{q}}_{i}\tilde{q}_{i}\right)\right.\nonumber\\
 && \left.+n\left(\frac{\hat{Q}_{r}Q_{r}}{2}+\frac{\hat{Q}_{i}Q_{i}}{2}\right)+\frac{n(n-1)}{2}\left(\hat{q}_{r}q_{r}+\hat{q}_{i}q_{i}\right)\right)\\
 &=& -\left(\hat{R}R+\frac{\left(\hat{Q}_{r}Q_{r}+\hat{Q}_{i}Q_{i}\right)}{2}-\frac{1}{2}\left(\hat{q}_{r}q_{r}+\hat{q}_{i}q_{i}\right)\right)
\end{eqnarray}

In order to recover the optimisation problems entailed in the curriculum procedure, we now have to consider the zero temperature limit of this expression. When $\beta\to\infty$, the order parameters follow non-trivial scaling laws:
\begin{itemize}
\item $\hat{Q}\to\beta^{2}\hat{Q}+\mathcal{O}\left(\beta\right)$, $\hat{q}\to\beta^{2}\hat{Q}$
\item $(\hat{Q}-\hat{q})\to-\beta\delta\hat{Q}$
\item $\hat{R}\to\beta\hat{R}$
\item $Q-q=\delta Q/\beta$
\end{itemize}
and similarly for the tilde parameters. Intuitively, looking at the last scaling law, we see that as the measure gets focused on the single minimiser of the loss, the overlap between different replicas $q$ rapidly converges to the norm $Q$. Moreover, the scaling with the inverse temperature of the conjugate parameters prevents the interaction term from becoming sub-dominant in the saddle-point. If we substitute the rescaled parameters in the above expression we obtain:
\begin{equation}
g_{i}=-\beta\left(\hat{R}R+\frac{1}{2}\left(\hat{Q}_{r}\delta Q_{r}-\delta\hat{Q}_{r}Q_{r}\right)+\frac{1}{2}\left(\hat{Q}_{i}\delta Q_{i}-\delta\hat{Q}_{i}Q_{i}\right)\right)
\end{equation}

\subsubsection*{Entropic term}

We can now compute a similar quantity for the entropic potential,  $g_i=\frac{\lim_{n\to 0}}{n}\log G_{S}\left(\tilde{R},R,\tilde{Q},Q\right)$. The general expression we will obtain can be specialised to the two cases $\left(\left\{ \tilde{R},R,\tilde{Q}_{r},Q_{r}\right\} ,\left\{ 0,0,\tilde{Q}_{i},Q_{i}\right\} \right)$ appearing in the free-entropy. After substituting the RS ansatz we find:
\small
\begin{eqnarray}
g_{S} &=& \lim_{\tilde{n}\to0}\frac{1}{n}\log\int\prod_{c}\left[d\tilde{W}^{c}e^{-\frac{\beta\gamma}{2}\left(\tilde{W}^{c}\right){}^{2}}\right]e^{-\frac{n\beta\gamma_{12}}{2}(\tilde{W}^{1})^{2}}\int\prod_{a}\left[dW^{a}e^{-\frac{\beta\left(\gamma+\gamma_{12}\right)}{2}\left(W^{a}\right){}^{2}}\right]\\
 &  & \times\exp\left(\hat{\tilde{R}}\sum_{c}\tilde{W}^{c}+\hat{R}\sum_{a}W^{a}+\frac{1}{2}\left(\hat{\tilde{Q}}-\hat{\tilde{q}}\right)\sum_{c}\left(\tilde{W}^{c}\right)^{2}+\frac{\hat{\tilde{q}}}{2}\left(\sum_{c}\tilde{W}^{c}\right)^{2}+\right.\nonumber\\
 &  & \left.+\frac{1}{2}\left(\hat{Q}-\hat{q}\right)\sum_{a}\left(W^{a}\right)^{2}+\frac{\hat{q}}{2}\left(\sum_{a}W^{a}\right)^{2}+\beta\gamma_{12}\sum_{a}\tilde{W}^{1}W^{a}\right)\nonumber\\
 & = & \lim_{\tilde{n}\to0}\frac{1}{n}\log\int\mathcal{D}z\int\mathcal{D}\tilde{z}\int\prod_{c}\left[d\tilde{W}^{c}e^{-\frac{\beta\gamma}{2}\left(\tilde{W}^{c}\right){}^{2}}\right]e^{-\frac{n\beta\gamma_{12}}{2}(\tilde{W}^{1})^{2}}\int\prod_{a}dW^{a}e^{-\frac{\beta\left(\gamma+\gamma_{12}\right)}{2}\left(W^{a}\right){}^{2}}\nonumber\\
 &  & \times\exp\left(\frac{1}{2}\left(\hat{\tilde{Q}}-\hat{\tilde{q}}\right)\sum_{c}\left(\tilde{W}^{c}\right)^{2}+\frac{1}{2}\left(\hat{Q}-\hat{q}\right)\sum_{a}\left(W^{a}\right)^{2}+\right.\nonumber\\
 &  & +\left.\left(\hat{\tilde{R}}+\sqrt{\hat{\tilde{q}}}\tilde{z}\right)\sum_{c}\tilde{W}^{c}+\left(\hat{R}+\beta\gamma_{12}W_{1}+\sqrt{\hat{q}}z\right)\sum_{a}W^{a}\right)\nonumber
 \end{eqnarray}
\begin{equation}
 = \int\mathcal{D}z\int\mathcal{D}\tilde{z}\frac{\int d\tilde{W}e^{-\frac{1}{2}(\beta\tilde{\gamma}-(\hat{\tilde{Q}}-\hat{\tilde{q}}))\tilde{W}^{2}+(\hat{\tilde{R}}+\sqrt{\hat{\tilde{q}}}\tilde{z})\tilde{W}}\,\log\left(\int dW\,e^{-\frac{1}{2}\left(\beta\left(\gamma+\gamma_{12}\right)-(\hat{Q}-\hat{q})\right)W^{2}+\left(\hat{R}+\beta\gamma_{12}W_{1}+\sqrt{\hat{q}}z\right)W}\right)}{\int d\tilde{W}e^{-\frac{1}{2}(\beta\tilde{\gamma}-(\hat{\tilde{Q}}-\hat{\tilde{q}}))\tilde{W}^{2}+(\hat{\tilde{R}}+\sqrt{\hat{\tilde{q}}}\tilde{z})\tilde{W}}}
\end{equation}
\normalsize

In the zero-temperature limit, we consider the same rescaling of the order parameters we described above. The integrals over the weights become an extremum operation:
\begin{equation}
g_{s}=\lim_{\beta\to\infty}\beta\int\mathcal{D}z\int\mathcal{D}\tilde{z}M_{s}^{\star},
\end{equation}
where:
\begin{equation}
M_{s}^{\star}=\mathrm{max}_{W}\left\{ -\frac{1}{2}\left(\left(\gamma+\gamma_{12}\right)+\delta\hat{Q}\right)W^{2}+\left(\hat{R}+\gamma_{12}\tilde{W}^{\star}+\sqrt{\hat{Q}}z\right)W\right\}
\end{equation}
\begin{equation}
=\frac{1}{2}\frac{\left(\hat{R}+\gamma_{12}\tilde{W}^{\star}+\sqrt{\hat{Q}}z\right)^{2}}{\left(\gamma+\gamma_{12}\right)+\delta\hat{Q}}
\end{equation}

and where: $\tilde{W}^{\star}=\mathrm{argmax}_{\tilde{W}}\left\{ -\frac{1}{2}(\tilde{\gamma}+\,\delta\hat{\tilde{Q}})\tilde{W}^{2}+(\hat{\tilde{R}}+\sqrt{\hat{\tilde{Q}}}\tilde{z})\tilde{W}\right\} =\frac{\hat{\tilde{R}}+\sqrt{\hat{\tilde{Q}}}\tilde{z}}{\tilde{\gamma}+\delta\hat{\tilde{Q}}}$.

Finally also the $\int\mathcal{D}z\int\mathcal{D}\tilde{z}$ integrations
can be carried out, giving:
\begin{equation}
\beta\int\mathcal{D}z\int\mathcal{D}\tilde{z}M_{s}^{\star}=\beta\int\mathcal{D}z\int\mathcal{D}\tilde{z}\frac{1}{2}\frac{\left(\hat{R}+\gamma_{12}\frac{\hat{\tilde{R}}+\sqrt{\hat{\tilde{Q}}}\tilde{z}}{\tilde{\gamma}+\,\delta\hat{\tilde{Q}}}+\sqrt{\hat{Q}}z\right)^{2}}{\left(\gamma+\gamma_{12}\right)+\delta\hat{Q}}
\end{equation}

\begin{equation}
=\frac{\beta}{2}\frac{\left(\hat{R}+\hat{\tilde{R}}\frac{\gamma_{12}}{\tilde{\gamma}+\delta\hat{\tilde{Q}}}\right)^{2}+\left(\frac{\gamma_{12}\sqrt{\hat{\tilde{Q}}}}{\tilde{\gamma}+\delta\hat{\tilde{Q}}}\right)^{2}+\hat{Q}}{\left(\gamma+\gamma_{12}\right)+\delta\hat{Q}}
\end{equation}

So, specialising to the the two terms that appear in the free-entropy we get:
\begin{equation}
g_S(\gamma_1,\gamma_2,\gamma_{12})=\rho\,g_{s}\left(\tilde{R},R,\tilde{Q}_{r},Q_{r}\right)+\left(1-\rho\right)\,g_{s}\left(0,0,\tilde{Q}_{i},Q_{i}\right) \label{eq:gs}
\end{equation}
\[
=\frac{\beta}{2}\left(\rho\frac{\left(\hat{R}+\hat{\tilde{R}}\frac{\gamma_{12}}{\gamma_1+\delta\hat{\tilde{Q}}_{r}}\right)^{2}+\left(\frac{\gamma_{12}\sqrt{\hat{\tilde{Q}}}_{r}}{\gamma_1+\delta\hat{\tilde{Q}}_{r}}\right)^{2}+\hat{Q}_{r}}{\left(\gamma_2+\gamma_{12}\right)+\delta\hat{Q}_{r}}+\left(1-\rho\right)\frac{\left(\frac{\gamma_{12}\sqrt{\hat{\tilde{Q}}_{i}}}{\gamma_1+\delta\hat{\tilde{Q}}_{i}}\right)^{2}+\hat{Q}_{i}}{\left(\gamma_2+\gamma_{12}\right)+\delta\hat{Q}_{i}}\right)
\]

\subsubsection*{Energetic term}
Since one of the two energetic terms appearing in the replicated free-energy depends on the $\tilde{n}$ replicas of the first weight configuration, and there is no interaction, we can take the $\tilde{n}\to0$ limit directly. Therefore we only have to evaluate the other contribution (dependent on the $n$ replicas of the second weight configuration). Defining $Q=Q_{r}+\Delta Q_{i}$, $Q=Q_{r}+\Delta Q_{i}$, we evaluate $g_E=\lim_{n\to0}\frac{1}{n}\log(G_E)$ in the RS ansatz:
\begin{equation}
g_E=\lim_{n\to0}\frac{1}{n}\log\int\frac{dud\hat{u}}{2\pi}e^{iu\hat{u}}e^{-\frac{\rho}{2}\left(\hat{u}\right)^{2}}\int\prod_{a}\left(\frac{d\lambda^{a}d\hat{\lambda}^{a}}{2\pi}e^{i\lambda^{a}\hat{\lambda}^{a}}\right)
\end{equation}
\[ 
\times e^{-\frac{1}{2}\left(Q-q\right)\sum_{a}\left(\hat{\lambda}_{a}\right)^{2}-\frac{1}{2}q\left(\sum_{a}\hat{\lambda}_{a}\right)^{2}-\hat{u}R\sum_{a}\hat{\lambda}_{a}-\beta\sum_{a}\ell\left(u,\lambda^{a}\right)}
\]
\begin{equation}
=\lim_{n\to0}\frac{1}{n}\log\int\frac{du}{\sqrt{2\pi\rho}}\int\prod_{a}\left(\frac{d\lambda^{a}d\hat{\lambda}^{a}}{2\pi}e^{i\lambda^{a}\hat{\lambda}^{a}}\right)
\end{equation}
\[ 
\times e^{-\frac{1}{2}\left(Q-q\right)\sum_{a}\left(\hat{\lambda}_{a}\right)^{2}-\frac{1}{2}q\left(\sum_{a}\hat{\lambda}_{a}\right)^{2}-\beta\sum_{a}\ell\left(u,\lambda^{a}\right)-\frac{1}{2\rho}\left(u+i\,R\sum_{a}\hat{\lambda}_{a}\right)^{2}}
\]
\begin{equation}
=\lim_{n\to0}\frac{1}{n}\log\int\mathcal{D}z_{0}\int\mathcal{D}u\left\{ \int\mathcal{D}\lambda e^{-\beta\,\ell\left(\sqrt{\rho}\,u,\sigma\left(\sqrt{\left(Q-q\right)}\lambda+\sqrt{q-\frac{m^{2}}{\rho}}z_{0}+\frac{m}{\sqrt{\rho}}u\right)\right)}\right\}^{n}
\end{equation}
\[
=\int\mathcal{D}z_{0}\int\mathcal{D}u\log\int\mathcal{D}\lambda e^{-\beta\,\ell\left(\sqrt{\rho}\,u,\sigma\left(\sqrt{Q-q}\lambda+\sqrt{q-\frac{m^{2}}{\rho}}z_{0}+\frac{m}{\sqrt{\rho}}u\right)\right)}
\]
So in the $\beta\to\infty$ limit, with the proper rescalings, we get:
\begin{equation}
g_{E}=\beta\int\mathcal{D}z\int\mathcal{D}u\,M_{E}^{\star}, \label{eq:ge}
\end{equation}
where:
\begin{equation}
M_{E}^{\star}=\max_{\lambda}-\frac{\lambda^{2}}{2}-\ell\left(\mathrm{sign}\left(\sqrt{\rho}\,u\right),\sigma\left(\sqrt{\delta Q_{r}+\Delta\delta Q_{i}}\lambda+\sqrt{Q_{r}+\Delta Q_{i}-\frac{R^{2}}{\rho}}z+\frac{R}{\sqrt{\rho}}u\right)\right)
\end{equation}

\subsection*{RS Free-entropy}
Finally, assuming the we can write down the RS free-entropy for the curriculum ordering as:
\begin{equation}
\Phi/\beta=-\mathrm{extr}\left(\hat{R}R+\frac{1}{2}\left(\left(\hat{Q}\delta Q-\delta\hat{Q}Q\right)_{r}+\left(\hat{Q}\delta Q-\delta\hat{Q}Q\right)_{i}\right)\right)
\end{equation}
\[ 
+\,g_{S}(\gamma_1,\gamma_2,\gamma_{12})+\alpha_{2}\,g_{E}\left(\Delta_2\right),
\]
where $g_S$ is defined in equation \eqref{eq:gs} and $g_E$ is defined in equation \eqref{eq:ge}.
The order parameters for the teacher system are obtained
independently from identical equations, after substituting $\lambda_1\to0, \lambda_2=\to\lambda_1$ and $\lambda_{12}\to0$, $\alpha_2\to\alpha_1$ and $\Delta_2\to\Delta_1$, and after adding a tilde to the remaining parameters. 

The saddle-point equations, yielding at convergence the asymptotic prediction for the order parameters, can be found by posing stationarity conditions for the free-entropy with respect to all overlaps. 

Note that, if instead of the simple setting just considered, where the data slice in the second stage has homogeneous variance for the irrelevant components, there are multiple subsets with different sizes and variances, the only variation in the free-entropy is in the energetic contribution. In general one will have a sum:
\begin{equation}
    \sum_s \alpha_s g_E(\Delta_s)
\end{equation}
over each of these subsets.

Moreover, if instead of two stages we consider multiple learning stages, the free-entropy for each successive step has an identical form, and one only has to substitute the tilde parameters with the order parameters obtained at the previous step. Note that the simplicity of nesting stages in this problem is connected to the convexity of this learning setting. Generally, adding more steps would increase the complexity of the calculation considerably.

\subsection*{Generalisation error}
With the saddle-point values for the order parameters, one can easily evaluate the generalisation error on new datapoints, which is the measure of performance we are employing in the main.
This performance can be obtained as:
\begin{equation}
1-\epsilon_{g}=\left\langle \Theta\left(\left(\frac{\pmb W_T\cdot x}{\sqrt{N}}\right)\left(\frac{\pmb W _2\cdot x}{\sqrt{N}}\right)\right)\right\rangle _{x(\Delta)}
\end{equation}
where $\Delta$ is the variance of the irrelevant components for the new pattern. A shortcut for evaluating this expression is to insert the order parameters in the expression through Dirac's $\delta$s. After a straightforward calculation, along the same lines of the one presented above, one obtains:
\begin{equation}
\epsilon_{g}=\frac{1}{\pi}\mathrm{arccos}\left(\frac{R}{\sqrt{\rho\left(Q_{r}+\Delta Q_{i}\right)}}\right).
\end{equation}
Of course, the generalisation accuracy is just the complementary quantity $1-\epsilon_g$.

\section{Additional results on sparsity}\label{si:sparsity}

\begin{figure}
	\centering
	\begin{subfigure}[b]{.32\linewidth}
        \centering
        \includegraphics[width=\linewidth]{img/batch-rhoVSalpha1-eta5.0_1Delta0.0_2Delta1.0-curriculum_vs_nocurriculum-optall-alphatot-coup.pdf}
		\caption{Curriculum learning.}
    \end{subfigure}
    \centering
	\begin{subfigure}[b]{.32\linewidth}
        \centering
        \includegraphics[width=\linewidth]{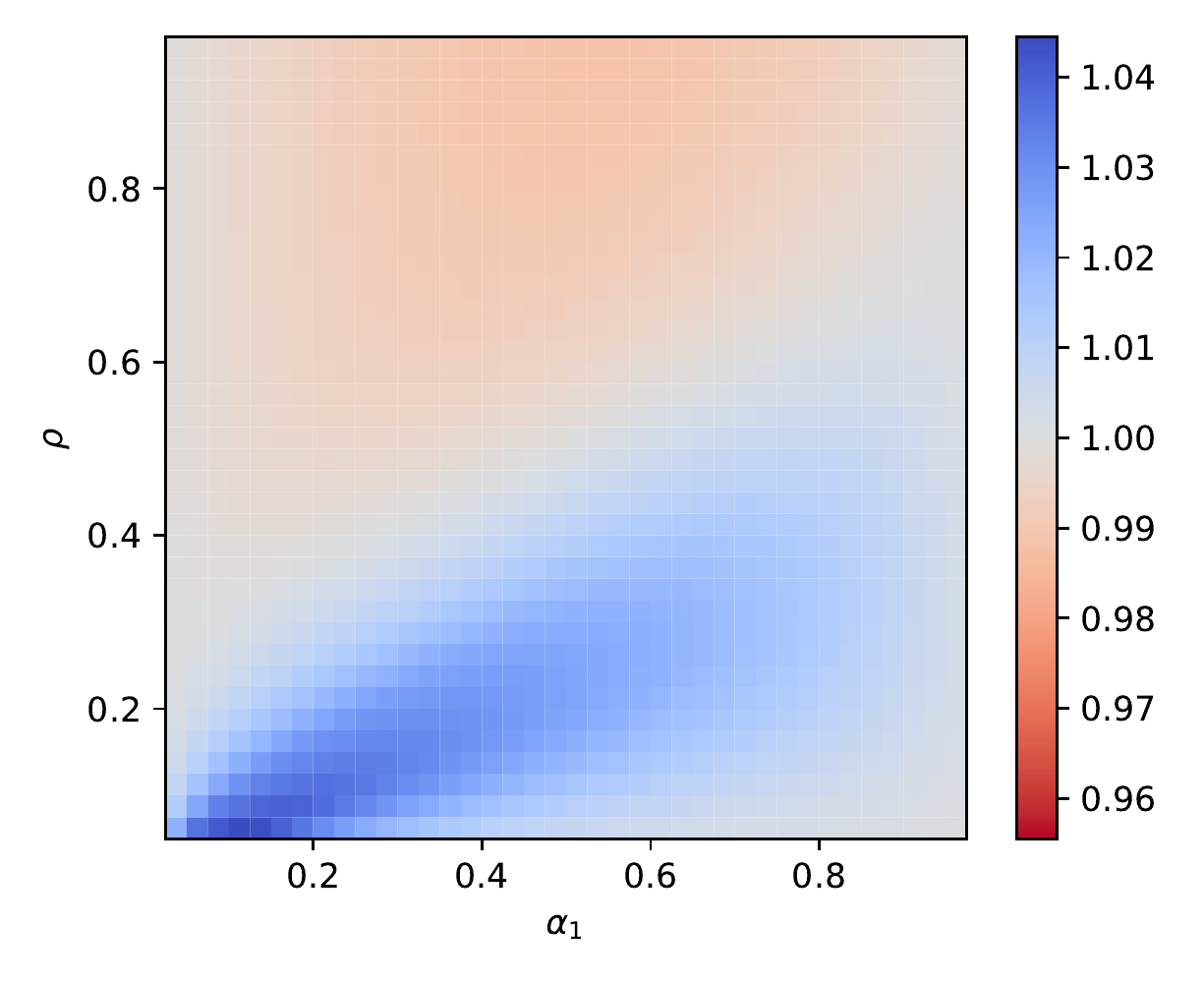}
		\caption{Anti-curriculum learning.}
    \end{subfigure}
	\begin{subfigure}[b]{.32\linewidth}
        \centering
        \includegraphics[width=\linewidth]{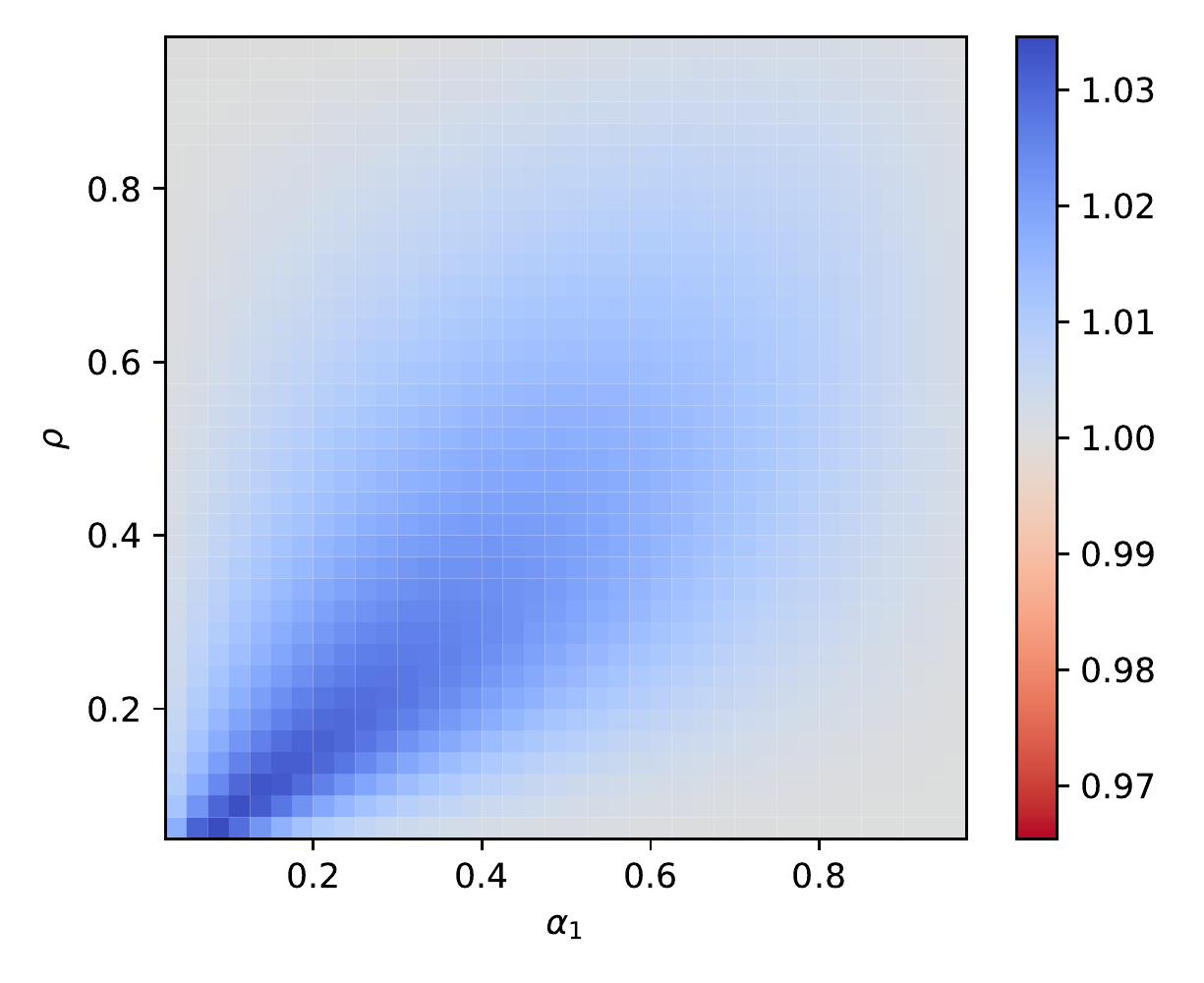}
		\caption{Curriculum vs anti-curriculum.}
    \end{subfigure}
    \caption{
    \textbf{Effect of sparsity.} Phase diagram on the effect of sparsity, Fig.~\ref{fig:coupling_effect}b, extended for all learning protocols.
    }
\end{figure}

We complement the discussion on the importance of sparsity, Sec.~\ref{sec:sparsity}, with the comparison with other learning protocols. Observe that anti-curriculum suffers the same issue of the curriculum method for sufficiently large fractions of relevant features $\rho$. In that regime, the splitting becomes sub-optimal because the solution found in the splitting does not provide enough information to help the other phase of learning. Consequently, the network is forced to set neglect the information in the batch in favour of exploring solutions further away from that one. This is outperform by standard learning, where all the bits of information are used.  

\section{Simulations on CIFAR10}
\label{sec:cifar}
\textbf{Task design.} Because a sparse set of relevant features is crucial to observing curriculum effects in our model, we created a task based on real data that has this property. In particular we create $32 \times 64$ pixel input examples by concatenating two images side-by-side from the CIFAR10 dataset. The correct output label is given by the label of the image on the left, while the image on the right is an irrelevant distractor. To vary difficulty, we scale the contrast of the irrelevant image. This dataset is meant to instantiate a simple example of learning an object classification amidst clutter. We emphasise that, as in our synthetic data model, each training sample always contains the same relevant and distractor images (i.e., we are not considering a data augmentation setting where each relevant image appears with many non-relevant images). To ensure no cross-contamination of training and testing samples, the distractor images for the training and test sets are drawn only from the same set.

\textbf{Model architecture and training regime.}
We train a single layer network with cross entropy loss (i.e. softmax regression), implemented in Pytorch Lightning by modifying the MIT-licensed \texttt{PyTorch\_CIFAR10} repository (\url{https://zenodo.org/record/4431043#.YLmz6zZKhsA}) to ensure that training parameters accord with standard practice. Networks were trained with SGD and Nesterov momentum, under default parameters: a learning rate of $1e-2$, momentum parameter 0.9,  batch size 256, and 100 epochs. The learning rate was annealed according to the `WarmUpCosine' schedule used in \texttt{PyTorch\_CIFAR10}, which linearly reduces the learning rate over the first $30\%$ of training steps before switching to a cosine shaped schedule on the remainder. 

\textbf{Experiment details and hyperparameter optimisation.} For the first phase of training, we used dataset sizes in 10 equal steps between 1000 and 50000. For the second phase, we used nine dataset sizes in 9 equal steps between 5333 and 48000. We optimised hyperparameters in each phase separately. In the first phase, we evaluated all combinations of initialisation scales of $\{0, .2, .5, 1.\}$, weight decay parameters of $\{0, .2, .5, 1., 2.\}$, and curriculum policy, for five random seeds. In the second phase, for each random seed and curriculum condition, we continued training from the best-performing model obtained in the first phase. We trained all combinations of five elastic penalties log spaced between $1e-3$ and $1e2$, and weight decay parameters $\{0, .2, .5\}$. We then compute the best performing model for each seed and take the mean over seeds. Finally, to evaluate the no-curriculum performance, we train shuffled dataset models with initialisation scales $\{0, .2, .5, 1.\}$ and weight decay parameters $\{0, .2, .5\}$. For visualisation purposes, we used nearest-neighbors interpolation in the phase portrait to provide values for all points used in the synthetic experiments. Experiments were run on V100 GPUs and required approximately 10000 GPU hours (including debugging and development), or $\approx 1110$ kg CO$_2$ eq according to the \href{https://mlco2.github.io/impact#compute}{MachineLearning Impact calculator} of Lacoste et al., 2019.

\section{Speed-up theory vs simulations}\label{sm:speedup}

As remarked in the main text, one of the advantages of the theoretical analysis is a huge speed-up in the time to collect the results, without need of averaging to reduce the fluctuations. 
In this section, we briefly report a comparison between the time required for the lines from theory and simulations shown in the main text.

In order to obtain figure 1c, a single run of the ODE equations takes ~2 milliseconds and a run of the simulations takes ~500 milliseconds. The figure is however obtained optimizing over all the hyperparameters (learning rate, initialization, weight decay) totalling ~400 milliseconds for the analytical solution; while, due to noise, simulation results for a single set of hyperparameters requires averaging 5000 realizations totalling ~41 minutes. We note that we did the hyperparameter optimization only once using the theoretical framework and then used the optima in the simulations in order to save compute time. The best comparison should therefore be done for a fixed set of hyperparameters and gives ~2 milliseconds vs ~41 minutes.  Overall, the analytical solution is between 2 and 6 orders of magnitude faster.



%% file: arxiv.bbl
\begin{thebibliography}{10}

\bibitem{lawrence1952transfer}
Douglas~H Lawrence.
\newblock The transfer of a discrimination along a continuum.
\newblock {\em Journal of Comparative and Physiological Psychology}, 45(6):511,
  1952.

\bibitem{baker1954discrimination}
Robert~A Baker and Stanley~W Osgood.
\newblock Discrimination transfer along a pitch continuum.
\newblock {\em Journal of Experimental Psychology}, 48(4):241, 1954.

\bibitem{elio_effects_1984}
Renee Elio and John Anderson.
\newblock The effects of information order and learning mode on schema
  abstraction.
\newblock {\em Memory \& Cognition}, 12:20--30, January 1984.

\bibitem{wilson_eighty_2019}
Robert~C. Wilson, Amitai Shenhav, Mark Straccia, and Jonathan~D. Cohen.
\newblock The {Eighty} {Five} {Percent} {Rule} for optimal learning.
\newblock {\em Nature Communications}, 10(1):4646, November 2019.

\bibitem{avrahami_teaching_1997}
Judith Avrahami, Yaakov Kareev, Yonatan Bogot, Ruth Caspi, Salomka Dunaevsky,
  and Sharon Lerner.
\newblock Teaching by {Examples}: {Implications} for the {Process} of
  {Category} {Acquisition}.
\newblock {\em The Quarterly Journal of Experimental Psychology Section A},
  50(3):586--606, August 1997.
\newblock Publisher: SAGE Publications.

\bibitem{pashler_when_2013}
Harold Pashler and Michael~C. Mozer.
\newblock When does fading enhance perceptual category learning?
\newblock {\em Journal of Experimental Psychology: Learning, Memory, and
  Cognition}, 39(4):1162--1173, 2013.

\bibitem{hornsby_improved_2014}
Adam~N. Hornsby and Bradley~C. Love.
\newblock Improved classification of mammograms following idealized training.
\newblock {\em Journal of Applied Research in Memory and Cognition},
  3(2):72--76, June 2014.

\bibitem{roads_easy--hard_2018}
Brett~D. Roads, Buyun Xu, June~K. Robinson, and James~W. Tanaka.
\newblock The easy-to-hard training advantage with real-world medical images.
\newblock {\em Cognitive Research: Principles and Implications}, 3, October
  2018.

\bibitem{the_international_brain_laboratory_standardized_2021}
{The International Brain Laboratory}, Valeria Aguillon-Rodriguez, Dora
  Angelaki, Hannah Bayer, Niccolo Bonacchi, Matteo Carandini, Fanny Cazettes,
  Gaelle Chapuis, Anne~K Churchland, Yang Dan, Eric Dewitt, Mayo Faulkner,
  Hamish Forrest, Laura Haetzel, Michael Häusser, Sonja~B Hofer, Fei Hu, Anup
  Khanal, Christopher Krasniak, Ines Laranjeira, Zachary~F Mainen, Guido
  Meijer, Nathaniel~J Miska, Thomas~D Mrsic-Flogel, Masayoshi Murakami,
  Jean-Paul Noel, Alejandro Pan-Vazquez, Cyrille Rossant, Joshua Sanders,
  Karolina Socha, Rebecca Terry, Anne~E Urai, Hernando Vergara, Miles Wells,
  Christian~J Wilson, Ilana~B Witten, Lauren~E Wool, and Anthony~M Zador.
\newblock Standardized and reproducible measurement of decision-making in mice.
\newblock {\em eLife}, 10:e63711, May 2021.
\newblock Publisher: eLife Sciences Publications, Ltd.

\bibitem{elman_learning_1993}
Jeffrey~L. Elman.
\newblock Learning and development in neural networks: the importance of
  starting small.
\newblock {\em Cognition}, 48(1):71--99, July 1993.

\bibitem{krueger_flexible_2009}
Kai~A. Krueger and Peter Dayan.
\newblock Flexible shaping: how learning in small steps helps.
\newblock {\em Cognition}, 110(3):380--394, March 2009.

\bibitem{bengio2009curriculum}
Yoshua Bengio, J{\'e}r{\^o}me Louradour, Ronan Collobert, and Jason Weston.
\newblock Curriculum learning.
\newblock In {\em Proceedings of the 26th annual international conference on
  machine learning}, pages 41--48, 2009.

\bibitem{pentina_curriculum_2015}
Anastasia Pentina, Viktoriia Sharmanska, and Christoph~H. Lampert.
\newblock Curriculum learning of multiple tasks.
\newblock pages 5492--5500. IEEE Computer Society, June 2015.
\newblock ISSN: 1063-6919.

\bibitem{HacohenW19}
Guy Hacohen and Daphna Weinshall.
\newblock On the power of curriculum learning in training deep networks.
\newblock In {\em ICML}, volume~97, pages 2535--2544. {PMLR}, 2019.

\bibitem{wu2020curricula}
Xiaoxia Wu, Ethan Dyer, and Behnam Neyshabur.
\newblock When do curricula work?
\newblock {\em ICLR}, 2020.

\bibitem{brown_language_2020}
Tom Brown, Benjamin Mann, Nick Ryder, Melanie Subbiah, Jared~D. Kaplan,
  Prafulla Dhariwal, Arvind Neelakantan, Pranav Shyam, Girish Sastry, Amanda
  Askell, Sandhini Agarwal, Ariel Herbert-Voss, Gretchen Krueger, Tom Henighan,
  Rewon Child, Aditya Ramesh, Daniel Ziegler, Jeffrey Wu, Clemens Winter, Chris
  Hesse, Mark Chen, Eric Sigler, Mateusz Litwin, Scott Gray, Benjamin Chess,
  Jack Clark, Christopher Berner, Sam McCandlish, Alec Radford, Ilya Sutskever,
  and Dario Amodei.
\newblock Language {Models} are {Few}-{Shot} {Learners}.
\newblock {\em Advances in Neural Information Processing Systems},
  33:1877--1901, 2020.

\bibitem{jiang_prioritized_2021}
Minqi Jiang, Edward Grefenstette, and Tim Rocktäschel.
\newblock Prioritized {Level} {Replay}.
\newblock {\em arXiv:2010.03934 [cs]}, January 2021.
\newblock arXiv: 2010.03934.

\bibitem{weinshall2018curriculum}
Daphna Weinshall, Gad Cohen, and Dan Amir.
\newblock Curriculum learning by transfer learning: Theory and experiments with
  deep networks.
\newblock In {\em International Conference on Machine Learning}, pages
  5238--5246. PMLR, 2018.

\bibitem{weinshall2020theory}
Daphna Weinshall and Dan Amir.
\newblock Theory of curriculum learning, with convex loss functions.
\newblock {\em Journal of Machine Learning Research}, 21(222):1--19, 2020.

\bibitem{ruiz2019tuning}
Miguel Ruiz-Garc{\'\i}a, Andrea~J Liu, and Eleni Katifori.
\newblock Tuning and jamming reduced to their minima.
\newblock {\em Physical Review E}, 100(5):052608, 2019.

\bibitem{mezard1987spin}
Marc M{\'e}zard, Giorgio Parisi, and Miguel~Angel Virasoro.
\newblock {\em Spin glass theory and beyond: An Introduction to the Replica
  Method and Its Applications}, volume~9.
\newblock World Scientific Publishing Company, 1987.

\bibitem{engel2001statistical}
Andreas Engel and Christian Van~den Broeck.
\newblock {\em Statistical mechanics of learning}.
\newblock Cambridge University Press, 2001.

\bibitem{zdeborova2016statistical}
Lenka Zdeborov{\'a} and Florent Krzakala.
\newblock Statistical physics of inference: Thresholds and algorithms.
\newblock {\em Advances in Physics}, 65(5):453--552, 2016.

\bibitem{bahri2020statistical}
Yasaman Bahri, Jonathan Kadmon, Jeffrey Pennington, Sam~S Schoenholz, Jascha
  Sohl-Dickstein, and Surya Ganguli.
\newblock Statistical mechanics of deep learning.
\newblock {\em Annual Review of Condensed Matter Physics}, 2020.

\bibitem{cugliandolo1993analytical}
Leticia~F Cugliandolo and Jorge Kurchan.
\newblock Analytical solution of the off-equilibrium dynamics of a long-range
  spin-glass model.
\newblock {\em Physical Review Letters}, 71(1):173, 1993.

\bibitem{biehl1995learning}
Michael Biehl and Holm Schwarze.
\newblock Learning by on-line gradient descent.
\newblock {\em Journal of Physics A: Mathematical and general}, 28(3):643,
  1995.

\bibitem{advani2020highdimensional}
M.S. Advani, A.M. Saxe, and H.~Sompolinsky.
\newblock High-dimensional dynamics of generalization error in neural networks.
\newblock {\em Neural Networks}, 132:428 -- 446, 2020.

\bibitem{goldt2019dynamics}
Sebastian Goldt, Madhu Advani, Andrew~M Saxe, Florent Krzakala, and Lenka
  Zdeborov\'{a}.
\newblock Dynamics of stochastic gradient descent for two-layer neural networks
  in the teacher-student setup.
\newblock In H.~Wallach, H.~Larochelle, A.~Beygelzimer, F.~d\textquotesingle
  Alch\'{e}-Buc, E.~Fox, and R.~Garnett, editors, {\em Advances in Neural
  Information Processing Systems}, volume~32. Curran Associates, Inc., 2019.

\bibitem{mannelli2019passed}
Stefano Sarao~Mannelli, Florent Krzakala, Pierfrancesco Urbani, and Lenka
  Zdeborova.
\newblock Passed \& spurious: Descent algorithms and local minima in spiked
  matrix-tensor models.
\newblock 97:4333--4342, 09--15 Jun 2019.

\bibitem{mannelli2019afraid}
Stefano~Sarao Mannelli, Giulio Biroli, Chiara Cammarota, Florent Krzakala, and
  Lenka Zdeborov{\'a}.
\newblock Who is afraid of big bad minima? analysis of gradient-flow in spiked
  matrix-tensor models.
\newblock In {\em Advances in Neural Information Processing Systems}, pages
  8679--8689, 2019.

\bibitem{mannelli2020complex}
Stefano Sarao~Mannelli, Giulio Biroli, Chiara Cammarota, Florent Krzakala,
  Pierfrancesco Urbani, and Lenka Zdeborov\'{a}.
\newblock Complex dynamics in simple neural networks: Understanding gradient
  flow in phase retrieval.
\newblock 33:3265--3274, 2020.

\bibitem{cui2020large}
Hugo Cui, Luca Saglietti, and Lenka Zdeborov{\'a}.
\newblock Large deviations for the perceptron model and consequences for active
  learning.
\newblock In {\em Mathematical and Scientific Machine Learning}, pages
  390--430. PMLR, 2020.

\bibitem{zenke2017continual}
Friedemann Zenke, Ben Poole, and Surya Ganguli.
\newblock Continual learning through synaptic intelligence.
\newblock In {\em International Conference on Machine Learning}, pages
  3987--3995. PMLR, 2017.

\bibitem{kirkpatrick2017overcoming}
James Kirkpatrick, Razvan Pascanu, Neil Rabinowitz, Joel Veness, Guillaume
  Desjardins, Andrei~A Rusu, Kieran Milan, John Quan, Tiago Ramalho, Agnieszka
  Grabska-Barwinska, et~al.
\newblock Overcoming catastrophic forgetting in neural networks.
\newblock {\em Proceedings of the national academy of sciences},
  114(13):3521--3526, 2017.

\bibitem{pavlov2010conditioned}
P~Ivan Pavlov.
\newblock Conditioned reflexes: an investigation of the physiological activity
  of the cerebral cortex.
\newblock {\em Annals of neurosciences}, 17(3):136, 2010.

\bibitem{skinner2019behavior}
Burrhus~Frederic Skinner.
\newblock {\em The behavior of organisms: An experimental analysis}.
\newblock BF Skinner Foundation, 2019.

\bibitem{ahissar1997task}
Merav Ahissar and Shaul Hochstein.
\newblock Task difficulty and the specificity of perceptual learning.
\newblock {\em Nature}, 387(6631):401--406, 1997.

\bibitem{morris1977levels}
C~Donald Morris, John~D Bransford, and Jeffery~J Franks.
\newblock Levels of processing versus transfer appropriate processing.
\newblock {\em Journal of verbal learning and verbal behavior}, 16(5):519--533,
  1977.

\bibitem{plunkett1991rote}
Kim Plunkett, Virginia Marchman, and Steen~Ladegaard Knudsen.
\newblock From rote learning to system building: acquiring verb morphology in
  children and connectionist nets.
\newblock In {\em Connectionist Models}, pages 201--219. Elsevier, 1991.

\bibitem{plunkett1991u}
Kim Plunkett and Virginia Marchman.
\newblock U-shaped learning and frequency effects in a multi-layered
  perception: Implications for child language acquisition.
\newblock {\em Cognition}, 38(1):43--102, 1991.

\bibitem{toneva2019empirical}
Mariya Toneva, Alessandro Sordoni, Remi~Tachet des Combes, Adam Trischler,
  Yoshua Bengio, and Geoffrey~J. Gordon.
\newblock An empirical study of example forgetting during deep neural network
  learning.
\newblock In {\em 7th International Conference on Learning Representations,
  {ICLR} 2019, New Orleans, LA, USA, May 6-9, 2019}. OpenReview.net, 2019.

\bibitem{saad1995exact}
David Saad and Sara~A Solla.
\newblock Exact solution for on-line learning in multilayer neural networks.
\newblock {\em Physical Review Letters}, 74(21):4337, 1995.

\bibitem{kocmi_curriculum_2017}
Tom Kocmi and Ondřej Bojar.
\newblock Curriculum {Learning} and {Minibatch} {Bucketing} in {Neural}
  {Machine} {Translation}.
\newblock In {\em Proceedings of the {International} {Conference} {Recent}
  {Advances} in {Natural} {Language} {Processing}, {RANLP} 2017}, pages
  379--386, Varna, Bulgaria, September 2017. INCOMA Ltd.

\bibitem{zhang2018empirical}
Nathan Schneider, Dirk Hovy, Anders Johannsen, and Marine Carpuat.
\newblock Semeval-2016 task 10: Detecting minimal semantic units and their
  meanings (dimsum).
\newblock In Steven Bethard, Daniel~M. Cer, Marine Carpuat, David Jurgens,
  Preslav Nakov, and Torsten Zesch, editors, {\em Proceedings of the 10th
  International Workshop on Semantic Evaluation, SemEval@NAACL-HLT 2016, San
  Diego, CA, USA, June 16-17, 2016}, pages 546--559. The Association for
  Computer Linguistics, 2016.

\bibitem{zhang_curriculum_2019}
Xuan Zhang, Pamela Shapiro, Gaurav Kumar, Paul McNamee, Marine Carpuat, and
  Kevin Duh.
\newblock Curriculum {Learning} for {Domain} {Adaptation} in {Neural} {Machine}
  {Translation}.
\newblock In {\em Proceedings of the 2019 {Conference} of the {North}
  {American} {Chapter} of the {Association} for {Computational} {Linguistics}:
  {Human} {Language} {Technologies}, {Volume} 1 ({Long} and {Short} {Papers})},
  pages 1903--1915, Minneapolis, Minnesota, June 2019. Association for
  Computational Linguistics.

\bibitem{pmlr-v70-zenke17a}
Friedemann Zenke, Ben Poole, and Surya Ganguli.
\newblock Continual learning through synaptic intelligence.
\newblock In Doina Precup and Yee~Whye Teh, editors, {\em Proceedings of the
  34th International Conference on Machine Learning}, volume~70 of {\em
  Proceedings of Machine Learning Research}, pages 3987--3995, International
  Convention Centre, Sydney, Australia, 06--11 Aug 2017. PMLR.

\bibitem{franz1997phase}
Silvio Franz and Giorgio Parisi.
\newblock Phase diagram of coupled glassy systems: A mean-field study.
\newblock {\em Physical review letters}, 79(13):2486, 1997.

\bibitem{saglietti2020solvable}
Luca Saglietti and Lenka Zdeborov{\'{a}}.
\newblock Solvable model for inheriting the regularization through knowledge
  distillation.
\newblock {\em CoRR}, abs/2012.00194, 2020.

\bibitem{clerkin_real-world_2017}
Elizabeth~M. Clerkin, Elizabeth Hart, James~M. Rehg, Chen Yu, and Linda~B.
  Smith.
\newblock Real-world visual statistics and infants' first-learned object names.
\newblock {\em Philosophical Transactions of the Royal Society of London.
  Series B, Biological Sciences}, 372(1711), January 2017.

\bibitem{liu2008easy}
Estella~H Liu, Eduardo Mercado~III, Barbara~A Church, and Itzel Ordu{\~n}a.
\newblock The easy-to-hard effect in human (homo sapiens) and rat (rattus
  norvegicus) auditory identification.
\newblock {\em Journal of Comparative Psychology}, 122(2):132, 2008.

\bibitem{kepple2022}
Daniel~R. Kepple, Rainer Engelken, and Rajan Kanaka.
\newblock Curriculum learning as a tool to uncover learning principles in the
  brain.
\newblock {\em ICLR}, 2022.

\bibitem{raz_how_2019}
Hadar~Karmazyn Raz, Drew~H. Abney, David Crandall, Chen Yu, and Linda~B. Smith.
\newblock How do infants start learning object names in a sea of clutter?
\newblock {\em Annual Conference of the Cognitive Science Society},
  2019:521--526, July 2019.

\bibitem{smith_developmental_2017}
Linda~B. Smith and Lauren~K. Slone.
\newblock A {Developmental} {Approach} to {Machine} {Learning}?
\newblock {\em Frontiers in Psychology}, 8, 2017.

\bibitem{yu_embodied_2012}
Chen Yu and Linda~B. Smith.
\newblock Embodied attention and word learning by toddlers.
\newblock {\em Cognition}, 125(2):244--262, November 2012.

\bibitem{krizhevsky_learning_nodate}
Alex Krizhevsky.
\newblock Learning {Multiple} {Layers} of {Features} from {Tiny} {Images}.
\newblock 2009.

\bibitem{orduna2012evoked}
Itzel Ordu{\~n}a, Estella~H Liu, Barbara~A Church, Ann~C Eddins, and Eduardo
  Mercado~III.
\newblock Evoked-potential changes following discrimination learning involving
  complex sounds.
\newblock {\em Clinical neurophysiology}, 123(4):711--719, 2012.

\bibitem{church2013temporal}
Barbara~A Church, Eduardo Mercado~III, Matthew~G Wisniewski, and Estella~H Liu.
\newblock Temporal dynamics in auditory perceptual learning: impact of
  sequencing and incidental learning.
\newblock {\em Journal of Experimental Psychology: Learning, Memory, and
  Cognition}, 39(1):270, 2013.

\bibitem{ruiz2021tilting}
Miguel Ruiz-Garcia, Ge~Zhang, Samuel~S Schoenholz, and Andrea~J Liu.
\newblock Tilting the playing field: Dynamical loss functions for machine
  learning.
\newblock In {\em International Conference on Machine Learning}, pages
  9157--9167. PMLR, 2021.

\end{thebibliography}
